\documentclass[abstract=true]{scrartcl}

\usepackage[english]{babel}

\usepackage{amsmath}
\usepackage{amssymb}
\usepackage{stmaryrd}
\usepackage{graphicx}
\usepackage{amsthm}

\usepackage{enumitem}
\usepackage{xifthen}

\renewcommand{\epsilon}{\varepsilon}

\newtheorem{lemma}{Lemma}[section]
\newtheorem{claim}[lemma]{Claim}
\newtheorem{theorem}[lemma]{Theorem}
\newtheorem{corollary}[lemma]{Corollary}

\newtheorem{proposition}[lemma]{Proposition}

\theoremstyle{definition}

\newtheorem{example}[lemma]{Example}
\newtheorem{definition}[lemma]{Definition}
\newtheorem{remark}[lemma]{Remark}
\newtheorem*{remark*}{Remark}

\newcommand{\var}[1]{\ensuremath{\operatorname{var}(#1)}}
\newcommand{\dvar}[1]{\ensuremath{\operatorname{dvar}(#1)}}
\newcommand{\dom}[1]{\ensuremath{\operatorname{dom}(#1)}}
\newcommand{\lit}[1]{\ensuremath{\operatorname{lit}(#1)}}

\newcommand\vek[1]{\mathbf{#1}}
\newcommand\assign[1]{\mathbf{#1}}

\newcommand{\litVar}[1]{\ensuremath{\llbracket #1\rrbracket}}

\renewcommand{\phi}{\varphi}

\newcommand{\amo}[1]{\ensuremath{\operatorname{\mathrm{AMO}}(#1)}}
\newcommand{\exone}[1]{\ensuremath{\operatorname{\mathrm{EO}}(#1)}}
\newcommand{\exoneN}{\ensuremath{\operatorname{\mathrm{EO}}}}
\newcommand{\Ds}{D^\mathrm{s}}
\newcommand{\Dc}{D^\mathrm{c}}

\newcommand{\amorep}[1]{\ensuremath{\operatorname{amo}(#1)}}
\newcommand{\amoenc}[1]{\ensuremath{\operatorname{amo}'(#1)}}
\newcommand{\amoencN}{\ensuremath{\operatorname{amo}'}}
\newcommand{\exonerep}[1]{\ensuremath{\operatorname{eo}(#1)}}
\newcommand{\exoneenc}[1]{\ensuremath{\operatorname{eo}'(#1)}}
\newcommand{\exoneencN}{\ensuremath{\operatorname{eo}'}}

\DeclareMathOperator{\Level}{Level}

\makeatletter 
\newcommand\mynobreakpar{\par\nobreak\@afterheading} 
\makeatother

\usepackage{array}
\newcounter{nnfgrprowcntr}[table]
\renewcommand{\thennfgrprowcntr}{N\arabic{nnfgrprowcntr}}
\newcolumntype{N}{>{\refstepcounter{nnfgrprowcntr}{\thennfgrprowcntr}}c}

\makeatletter
\newcommand{\manuallabel}[2][]{\ifthenelse{\isempty{#1}}{#2\def\@currentlabel{#2}\label{#2}}{#1\def\@currentlabel{#1}\label{#2}}}
\makeatother

\usepackage{booktabs}

\begin{document}

\title{Propagation complete encodings of smooth DNNF theories}
\author{Petr Ku{\v c}era\thanks{Department of Theoretical Computer Science
and Mathematical Logic,
Faculty of Mathematics and Physics,
Charles University, Czech Republic,
kucerap@ktiml.mff.cuni.cz}
   \and Petr Savick{\'y}\thanks{Institute of Computer Science,
   The Czech Academy of Sciences,
   Czech Republic,
   savicky@cs.cas.cz}
}
\date{\empty}

\maketitle

\begin{abstract}
   We investigate conjunctive normal form (CNF) encodings of a function represented
with a decomposable negation normal form (DNNF). Several
   encodings of DNNFs and decision diagrams were considered by
   (Ab{\'\i}o et al. 2016). The authors differentiate
   between encodings which implement consistency or domain consistency
by unit propagation from encodings which are unit refutation complete or
propagation complete. The difference is that in the former case we do not
   care about propagation strength of the encoding with respect to the auxiliary
   variables while in the latter case we treat all variables (the
   main and the auxiliary ones) in the same way.
The currently known encodings of
DNNF theories implement domain consistency. Building on these
encodings we generalize the result of (Ab{\'\i}o et al. 2016) on
a propagation complete encoding of decision diagrams and present
a propagation complete encoding of a DNNF and its generalization
for variables with finite domains.
\end{abstract}

\section{Introduction}

Decomposable negation normal forms (DNNFs) were introduced
in~\cite{D99} where an approach of compiling a conjunctive normal form
(CNF) into a DNNF was described. Since then DNNFs and their subclasses
were extensively studied as a target language in knowledge
compilation. DNNFs form the most succinct language within the knowledge
compilation map~\cite{DM02} that allows efficient consistency
checking, clause entailment and model enumeration queries. Specific
restrictions are often put on DNNFs in order to get efficient
answering of other queries as well. A prominent example is the
language of deterministic DNNFs (d-DNNFs) which additionally allow model counting.
For specific applications of DNNFs, see for
example~\cite{B04,EW06,H06,PBDG05,VAE11}.
Ordered binary decision diagrams (OBDD) and their
generalization in the form of multivalued decision diagrams (MDD) form
subclasses of d-DNNFs which are also often used in both theory and
applications.

If a constraint represented with an MDD, d-DNNF or a DNNF is a part
of a larger problem, efficient propagators for these
representations~\cite{GS12,GSS11,GSH13} can be used to maintain
domain consistency of the constraint within a constraint programming
solver. Recently, in~\cite{UGSS19} the authors have considered a
general problem of compiling global constraints into multivalued
decision diagrams (MDD) and d-DNNFs that can be used in this way.
Besides detecting conflicts, a propagator can generate an explanation,
which is a clause that can be added to the instance. This is the
basis for lazy clause generation strategy in CP solvers which
can increase their efficiency considerably and is analogous to the CDCL
strategy in SAT solvers, see, for example~\cite{BP20,GHS19}.

There are also approaches using an explicit CNF encoding of
a constraint in a CP solver. A CNF encoding can be included into
the instance at the beginning of the search or only if the constraint
appears to be active in the search process as in lazy
approaches~\cite{ANORS13}, where a propagator that participates in
many conflicts is replaced by a CNF decomposition during run-time.
A solver can also resolve a conflict by extending
the instance by clauses that contain auxiliary variables new in the
instance by lazily expanding a corresponding part of a large known
encoding of the constraint~\cite{AS12,GHS19}.
Specific situations, where a CNF encoding of a DNNF is suitable for
use in a SAT instance are described in~\cite{PG05,SS10a}.

The running time of a clause learning SAT solver can be exponentially
larger, if branching in the solver is restricted to make decisions
only on the input variables of an encoding, see~\cite{JJ09}.
A similar effect was observed in~\cite{ANORS13}, when a CP
solver cannot make decisions on auxiliary variables of an encoding
used for generating clauses by explaining conflicts in a propagator.
This suggests that the solver can benefit from branching on
auxiliary variables and then, it is natural to consider
propagation strength on all variables of the encoding and not only on
the main or input variables required by domain consistency.
Here, by propagation, we mean unit propagation which is a standard
tool in SAT solvers based on CDCL\@.

Let us recall several levels of propagation strength of a CNF
encoding of a constraint \(f(\vek{x})\).
For simplicity, we assume that the variables $\vek{x}$
have boolean domains although later we consider the direct
encoding (see for example~\cite{AGMES16,BKNW09}) to encode variables
with arbitrary finite domains. We say that a CNF formula
\(\varphi(\vek{x}, \vek{y})\) is a \emph{CNF encoding} of \(f(\vek{x})\)
if \(f(\vek{x})\equiv(\exists \vek{y})\varphi(\vek{x}, \vek{y})\).
The variables \(\vek{y}\) are called auxiliary variables and are used
only for the purpose of the encoding.
Following~\cite{AGMES16} we consider several properties of a CNF
encoding \(\varphi(\vek{x}, \vek{y})\) of a function \(f(\vek{x})\),
which specify its propagation strength.
In the description below,
a partial assignment is represented by a consistent
set of literals and implements means by unit propagation.

\begin{itemize}
   \item Encoding \(\varphi\) \emph{implements consistency}
      (called also a \emph{consistency checker} in~\cite{BKNW09}) if for
      any partial assignment \(\alpha\) to the main variables $\vek{x}$,
      we have that if \(f(\vek{x}) \wedge \alpha\) is inconsistent, then
      unit propagation derives a contradiction
      from \(\varphi(\vek{x}, \vek{y})\land\alpha\).
   \item Encoding \(\varphi\) \emph{implements domain consistency}
      (called  also \emph{generalized arc consistency (GAC)} or
      a \emph{propagator} in~\cite{BKNW09}) if
      for every partial assignment \(\alpha\) to the main variables $\vek{x}$
      and every literal \(l\) on a main variable, such that
      \(f(\vek{x}) \wedge \alpha \models l\), unit
      propagation on \(\varphi(\vek{x}, \vek{y})\land\alpha\) derives \(l\) or
      a contradiction.
   \item Encoding \(\varphi\) \emph{implements unit refutation
         completeness} (called also \emph{unit refutation complete} or
      \emph{URC} encoding, \cite{V94}) if for any partial assignment \(\alpha\) to
      the variables $\vek{x} \cup \vek{y}$, such that
      \(\phi(\vek{x}, \vek{y}) \wedge \alpha \models \bot\),
      unit propagation derives a contradiction from \(\varphi\land\alpha\).
   \item Encoding \(\varphi\) \emph{implements propagation
         completeness} (called also \emph{propagation complete} or
      \emph{PC} encoding, \cite{BM12}) if for any partial assignment \(\alpha\) to
      the variables $\vek{x} \cup \vek{y}$ and any literal $l$ on these
      variables, such that \(\phi(\vek{x}, \vek{y}) \wedge \alpha \models l\),
      unit propagation on \(\varphi\land\alpha\) derives \(l\) or
      a contradiction.
\end{itemize}

Clearly, a propagation complete encoding implements domain consistency
and a unit refutation complete encoding implements consistency. By~\cite{GK13},
the class of URC formulas coincides with the class SLUR introduced
in~\cite{AFSS95}.
Encodings which enforce GAC (domain consistency) by unit propagation were
considered e.g.~in~\cite{B07,BKNW09}.

Encoding of a DNNF that implements consistency is relatively straightforward,
since by decomposability propagating zeros from the inputs to the output is
sufficient for testing consistency of partial assignments of the inputs. The
basic idea how to obtain an encoding of a smooth DNNF of linear size implementing
domain consistency (enforcing GAC by unit propagation), appeared first
in~\cite{QW07} for a structure representing context free grammars.
Although the authors do not mention it, the structure is a special
case of a smooth DNNF and the construction of its encoding enforcing
domain consistency by unit propagation can be generalized to an arbitrary smooth
DNNF\@. This generalization is described in~\cite{BJKW08}, although
the necessary assumption of smoothness is not explicitly mentioned.
This assumption is used in~\cite{GS12} and~\cite{AGMES16} where the description
includes also the clauses that are needed, if some of the literals on
input variables do not appear in the DNNF\@.

In~\cite{AGMES16}, an encoding implementing propagation completeness
for a function given by an MDD is presented. A decision diagram is
a special case of a DNNF if we rewrite
each decision node with a disjunction of two conjunctions in the
standard way. The authors of~\cite{AGMES16} posed a question whether
a propagation complete encoding of a sentential decision diagram~\cite{D11}
or some more general representation within the class of NNFs can be
determined. In this paper, we present a polynomial time construction
of such an encoding for a general DNNF and its generalization
for variables with arbitrary finite domains which was also considered
in~\cite{GS12} and which we call
multivalued DNNF\@.
For technical reasons, the construction is formulated for
a smooth (multivalued) DNNF\@. Asymptotically, this is not a strong restriction,
since a DNNF can be transformed into a smooth DNNF with a polynomial
increase of size~\cite{D01a} and the same holds for a multivalued DNNF\@.
The presented encoding was used for a construction of a propagation complete
encoding for a related model more expressive than DNNF in~\cite{KS21}.

For our construction, we need an encoding of the at-most-one constraint
and the exactly-one constraint of linear size that can be used to replace
a prime representation of these constraints inside a larger encoding while
preserving the propagation strength of the whole encoding.
We show that the usual 2-CNF encodings of the at-most-one constraint and
the ladder encoding of the exactly-one constraint~\cite{GN04} satisfy this
property. On the other hand, we demonstrate that the encoding of exactly-one constraint obtained
by combining an encoding of at-most-one constraint with the disjunction of all
the variables is not suitable for this purpose.

In Section~\ref{sec:definitions}, we recall the necessary notions and
terminology.
Section~\ref{sub:fullnnf} demonstrates an
example of a smooth d-DNNF, for which the known encodings implementing
domain consistency are not propagation complete.
Section~\ref{sub:mdnnf} introduces multivalued DNNF as a generalization
of a DNNF for variables with finite domains, defines the notion of
a separator cover used in our construction, and
describes a URC and a PC encoding
of a multivalued DNNF which is the main result.
Sections~\ref{sub:mst} introduces the notions which
we use to analyze our encodings.
Section~\ref{sec:separators} describes a method how to obtain
a separator cover.
The proof of the main result is in Section~\ref{sec:encoding}.
In Section~\ref{sec:amo-eo}, we prove the properties of encodings of
the at-most-one and
the exactly-one constraint used in our construction
and estimate the size of the encoding of multivalued DNNF with encodings
of at-most-one and exactly-one of linear size.
Section~\ref{sec:conclusion}
formulates some questions for further research.



\section{Definitions}
\label{sec:definitions}

In this section, we recall the main notions used in our paper.

\subsection{Propagation and Unit Refutation Complete Encodings}
\label{sub:pc-enc} 

A formula in conjunctive normal form (\emph{CNF formula}) is
a conjunction of clauses. A \emph{clause} is a disjunction of a set of
literals and a \emph{literal} is a variable \(x\) (\emph{positive literal}) or its negation
\(\neg x\) (\emph{negative literal}). Given a set of variables \(\vek{x}\), \(\lit{\vek{x}}\)
denotes the set of literals on variables in \(\vek{x}\).

A \emph{partial assignment} \(\alpha\) of values to variables in \(\vek{x}\) is
a subset of \(\lit{\vek{x}}\) that does not contain a complementary
pair of literals, so we have \(|\alpha\cap\lit{x}|\le 1\) for each
\(x\in\vek{x}\). We identify a set of literals \(\alpha\) (in
particular a partial assignment) with the conjunction of these
literals if \(\alpha\) is used in a formula such as
\(\varphi(\vek{x})\land\alpha\).
A mapping \(\assign{a}:\vek{x}\to\{0, 1\}\) or, equivalently,
\(\assign{a}\in{\{0, 1\}}^\vek{x}\) represents
a \emph{full assignment} of values to \(\vek{x}\).
Alternatively, a full assignment can be represented with the set of literals
satisfied by the assignment. We use these representations interchangeably.

We consider encodings of boolean functions defined as follows.

\begin{definition}[Encoding]
   \label{def:cnf-enc} 
   Let \(f(\vek{x})\) be a boolean function on variables \(\vek{x}=(x_1, \dots,
      x_n)\). Let \(\varphi(\vek{x},\vek{y})\) be a CNF formula
   on \(n+m\) variables where \(\vek{y}=(y_1, \dots,
      y_m)\).
   We call \(\varphi\) a \emph{CNF encoding} of \(f\) if
   for every \(\assign{a}\in{\{0, 1\}}^{\vek{x}}\) we have
   \begin{equation}
      \label{eq:enc-def}
      f(\assign{a}) \equiv (\exists
      \assign{b}\in{\{0, 1\}}^{\vek{y}})\, \varphi(\assign{a},
      \assign{b}) \text{\,.}
   \end{equation}
The variables
   in \(\vek{x}\) and \(\vek{y}\) are called \emph{main variables} and
   \emph{auxiliary variables}, respectively.
\end{definition}

We are interested in encodings which are propagation complete or at
least unit refutation complete. These notions rely on
unit propagation which is a well known procedure in SAT
solving~\cite{BHMW09}. For technical reasons, we represent
unit propagation using the following two rules
\begin{equation} \label{eq:derive-a-literal}
l \vee l_1 \vee \cdots \vee l_k, \neg l_1, \ldots, \neg l_k \vdash l
\end{equation}
\begin{equation} \label{eq:derive-a-contradiction}
l, \neg l \vdash \bot \;.
\end{equation}
Derivation of a contradiction and derivation of the literals
in non-contradictory cases yields the same result using these rules
and using the unit propagation implemented in a SAT solver.
In contradictory cases, the set of the literals
derived together with a contradiction may be different.
We say that a literal $l$ or $\bot$ can be derived from
\(\varphi\) by \emph{unit propagation}
and denote this fact with \(\varphi\vdash_1 l\) or $\phi \vdash_1 \bot$,
respectively,
if the unit clause \(l\) or $\bot$, respectively,
is in $\phi$ or can be derived from
\(\varphi\) by a series of applications of
the rules~\eqref{eq:derive-a-literal} and~\eqref{eq:derive-a-contradiction}.

We say that a set of literals, in particular,
a partial assignment \(\alpha\) is
\emph{closed under unit propagation} in \(\varphi\),
if for every literal \(l\) we
have that \(\varphi\land \alpha\vdash_1 l\) implies \(l\in\alpha\).
The \emph{closure under unit propagation} of a set of literals $\alpha$
is the smallest superset of $\alpha$ that is closed under unit propagation.
If $\phi \wedge \alpha \not\vdash_1 \bot$, then $\alpha$ and its closure
are partial assignments.

The notion of a propagation complete CNF formula
was introduced in~\cite{BM12} as a strengthening of a unit refutation
complete CNF formula introduced in~\cite{V94}. These notions allow to
distinguish different levels of propagation strength depending on the
type of propagation (URC or PC) and the set of variables involved in
the propagation.

\begin{definition}
   \label{def:pc-enc} 
   Let \(\varphi(\vek{x}, \vek{y})\) be a CNF encoding of a boolean
   function defined on a set of variables \(\vek{x}\)
and let $\vek{v} \subseteq \vek{x} \cup \vek{y}$ be non-empty.
   \begin{itemize}
      \item We say that the encoding \(\varphi\) is \emph{unit refutation
            complete (URC) on the variables $\vek{v}$}, if the
         following implication holds for every partial assignment
         \(\alpha\subseteq\lit{\vek{v}}\)
         \begin{equation}
            \label{eq:urc}
            \varphi(\vek{x}, \vek{y})\land\alpha\models\bot\Longrightarrow\varphi(\vek{x}, \vek{y})\land\alpha\vdash_1\bot
         \end{equation}
      \item We say that the encoding \(\varphi\) is a \emph{propagation complete
            (PC) on the variables $\vek{v}$}, if for every partial assignment
         \(\alpha\subseteq\lit{\vek{v}}\) and every \(l\in\lit{\vek{v}}\), such that
         \begin{equation}
            \label{eq:pc-enc-1}
            \varphi(\vek{x}, \vek{y})\land\alpha\models l
         \end{equation}
         we have
         \begin{equation}
            \label{eq:pc-enc-2}
            \varphi(\vek{x}, \vek{y})\wedge\alpha\vdash_1 l
            \hspace{1em}\text{or}\hspace{1em}
            \varphi(\vek{x}, \vek{y})\wedge\alpha\vdash_1 \bot\text{.}
         \end{equation}
\item If an encoding is URC on the variables $\vek{v} = \vek{x}$, we say that
it \emph{implements consistency} (by unit propagation).
\item If an encoding is PC on the variables $\vek{v} = \vek{x}$, we say that
it \emph{implements domain consistency} (by unit propagation).
\item If an encoding is URC or PC on the variables $\vek{v} = \vek{x} \cup \vek{y}$,
we say that it is URC or PC, respectively.
   \end{itemize}
\end{definition}

Given a set of literals \(A\), we use \(\amo{A}\) to denote the
\emph{at-most-one} constraint which is satisfied if and only if at
most one of the literals in \(A\) is satisfied. If \(A\) is
specified within the specification of the constraint, we usually
drop the curly brackets (e.g.\ we write \(\amo{x_1, \neg x_2, x_3}\)
instead of \(\amo{\{x_1, \neg x_2, x_3\}}\)). 
We use \(\amorep{A}\) to denote the CNF representation of \(\amo{A}\)
which consists of all clauses \(\neg l_1\lor \neg l_2\),
\(l_1, l_2\in A\), \(l_1\neq l_2\).
Since \(\amorep{A}\) consists of all prime
implicates of \(\amo{A}\), it is propagation complete.

Similarly, we use \(\exone{A}\) to denote the
\emph{exactly-one} constraint which is satisfied if and only if
exactly one of the literals in \(A\) is satisfied. 
We use \(\exonerep{A}\) to denote the CNF representation of $\exone{A}$
consisting of \(\amorep{A}\) together with
the clause \(\bigvee_{l\in A}l\). Since \(\exonerep{A}\) consists of all prime
implicates of \(\exone{A}\), it is propagation complete.

\subsection{Decomposable Negation Normal Form}

The notion of a DNNF was introduced in~\cite{D99}
as a restricted NNF\@.
A sentence in \emph{negation normal form} (NNF) \(D\) is a rooted DAG with
vertices \(V\), root \(\rho\in V\), the set of edges \(E\), and the set
of leaves \(L\subseteq V\). The inner vertices in $V$ are labeled with
\(\land\) or \(\lor\) and they represent connectives or gates in a monotone
circuit. Each edge $(v,u)$ in \(D\) connects an inner vertex $v$ labeled
\(\land\) or \(\lor\) with one of its inputs $u$. The edge is directed
from $v$ to $u$, so the inputs of a vertex are its successors (or child
vertices). The leaves are labeled with constants \(0\) or \(1\), or
literals \(l\in\lit{\vek{x}}\) where \(\vek{x}\) is a set of
variables.

For a vertex \(v\in V\), let us denote \(\var{v}\) the set of variables
from $\vek{x}$ that appear in the leaves which can be reached from \(v\)
by a directed path.
More precisely, a variable \(x\in\vek{x}\) belongs to
\(\var{v}\) if there is a directed path from \(v\) to a leaf vertex labeled with
a literal from $\lit{x}$.

\begin{definition}[\cite{DM02}]\label{def:decomp-smooth-bool}
    We define the following structural restrictions of NNFs.\mynobreakpar%
   \begin{itemize}
      \item We say that NNF \(D\) is \emph{decomposable} (DNNF), if
         for every vertex \(v=u_1\land\dots\land u_k\) we have that the
         sets of variables \(\var{u_1}, \dots, \var{u_k}\) are
         pairwise disjoint.
      \item We say that DNNF \(D\) is \emph{smooth} if for every vertex
         \(v=u_1\lor\dots\lor u_k\) we have
         \(\var{v}=\var{u_1}=\dots=\var{u_k}\).
   \end{itemize}
\end{definition}

Decomposability is a strong restriction. In particular, the satisfiability
test is polynomial for a DNNF, while it is NP-complete for a general NNF\@.
On the other hand, it is possible to transform a general DNNF into an
equivalent smooth DNNF with a polynomial increase of size~\cite{D01a}.
For example, a simple disjunction $x_1 \vee x_2$ is a DNNF that is not
smooth. We can form an equivalent smooth DNNF by adding trivial
subformulas to obtain
$x_1 \wedge (x_2 \vee \neg x_2) \vee (x_1 \vee \neg x_1) \wedge x_2$.

\subsection{Propagation Strength of Encodings of DNNFs and MDDs} \label{sub:fullnnf}

The known encodings of DNNF implementing domain consistency
are different from Tseitin encoding
of a circuit in that the models of these encodings do not represent the exact
computations. The models are assignments of values to
gates which assign the value $1$ to the output gate and to some of
the gates satisfied in the computation. The gates set to $1$ are
chosen so that the model is sufficient to certify that the output is $1$,
however, in general, not all of the gates satisfied in the computation
are needed to certify the output.
This modification appeared to be useful
to construct encodings implementing domain consistency. Our PC encoding
goes further in this
direction in the sense that the models of the encoding are restricted
to the minimal certificates of satisfiability of the circuit which
is not the case of the previous encodings of DNNFs.

Let us point out that some modification of the set of models of
a Tseitin encoding is not only useful, but necessary to obtain
a URC encoding of a DNNF\@. It is a well-known fact that testing
satisfiability of a conjunction of two DNNFs is NP-complete. See,
for example, the proof of the fact that DNNFs do not satisfy bounded
conjunction closure in~\cite{DM02}. If $F(\vek{x})$ and $G(\vek{x})$
are DNNFs, then $F \vee G$ is also a DNNF and its Tseitin encoding
contains vertices $v_F$ and $v_G$ corresponding to the outputs of
$F$ and $G$. If an encoding $\phi$ of $F \vee G$ extends Tseitin
encoding by additional clauses and is URC, then
$\phi \wedge \alpha \wedge v_F \wedge v_G$, where $\alpha$
is a partial assignment of the main variables $\vek{x}$, should
derive a contradiction by unit propagation if and only if
$F(\vek{x}) \wedge G(\vek{x}) \wedge \alpha$ is unsatisfiable.
Unless P is equal to NP, this cannot be achieved by additional clauses
computable in polynomial time. On the other hand, the known encodings of DNNF
avoid this problem and can be extended to a URC or even a PC encoding
in polynomial time.
The encodings contain the variables $v_F$ and $v_G$, however,
they do not guarantee that both these variables are
satisfied in every model representing a satisfying computation
of $F(\vek{x}) \vee G(\vek{x})$ for an assignment of the main
variables $\vek{x}$ which additionally satisfies
$F(\vek{x}) \wedge G(\vek{x})$.

Let us briefly describe an example of a DNNF for which the \textsf{FullNNF}
encoding described in~\cite{AGMES16} is not PC.
Assume \(D\) is a smooth DNNF with vertices \(V\), root \(\rho\in V\),
the set of edges \(E\), and the set of leaves \(L\subseteq V\).
Let us assume that \(D\) represents boolean function \(f(\vek{x})\)
and the leaves of \(D\) are associated with the literals
\(\lit{\vek{x}}\).
\textsf{FullNNF} encoding of \(D\) is a CNF formula \(\psi(\vek{x},
   \vek{v})\) where \(\vek{v}\) are auxiliary variables corresponding
to the inner vertices of \(D\). It consists of clauses~\ref{FN1}
to~\ref{FN4} described in Table~\ref{tab:clauses:fullnnf} and the unit
clause \(\rho\).

\begin{table}[t]
   \begin{center}
      \begin{tabular}{l l l}
         \toprule
         \multicolumn{1}{c}{group} & clause & condition\\
         \midrule
         \manuallabel[N1]{FN1} 
         & \(v\to v_1\vee\dots\vee v_k\) &\(v=v_1\vee\dots\vee v_k\)\\
         \manuallabel[N2]{FN2} 
         & \(v\to v_i\) &\(v=v_1\land\dots\land v_k\), \(i=1, \dots, k\)\\
         \manuallabel[N3]{FN3} 
         & \(v\to p_1\vee\dots\vee p_k\) &\(v\) has incoming edges from
         \(p_1, \dots, p_k\)\\
         \manuallabel[N4]{FN4} 
         & $\neg l$ & $l \not\in L$\\
         \bottomrule
      \end{tabular}
   \end{center}
   \caption{Clauses of the \textsf{FullNNF} encoding.}\label{tab:clauses:fullnnf}
\end{table}

   \begin{figure}
      \begin{center}
         \includegraphics*{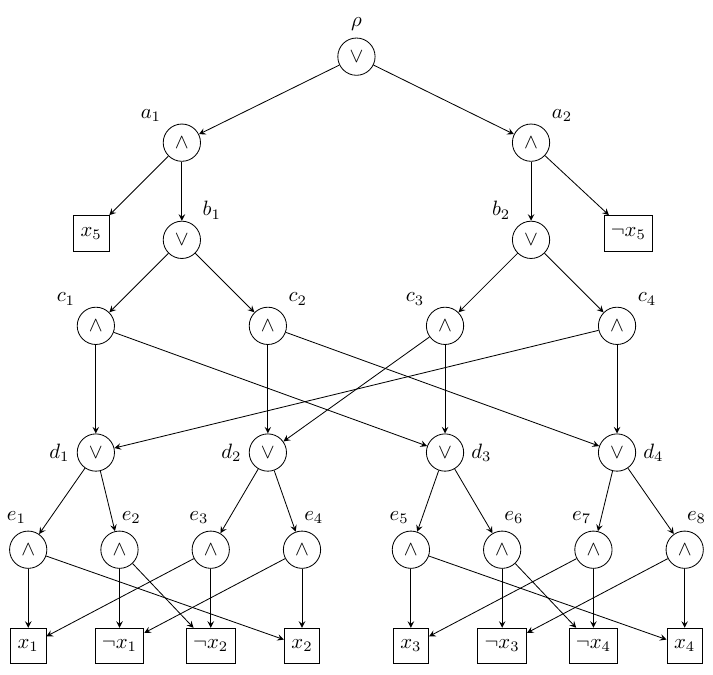}
      \end{center}
      \caption{An example DNNF\@. The root \(\rho\) together with vertices
         \(a_1\) and \(a_2\) represents a decision vertex on variable
         \(x_5\). Vertex labeled \(d_1\) represents
         disjunctive normal form (DNF)
         \(x_1 x_2 \lor\overline{x}_1\overline{x}_2\) (where we use the
         usual compressed form of conjunctions of literals) which is
         equivalent to condition \(x_1=x_2\).
         Vertex \(d_2\) represents DNF \(\overline{x}_1x_2\lor
            x_1\overline{x}_2\)
         which is equivalent to \(x_1\neq x_2\). Similarly, vertex \(d_3\)
         represents \(x_3x_4\lor \overline{x}_3\overline{x}_4\)
         (\(x_3=x_4\)) and vertex \(d_4\) represents
         \(x_3\overline{x}_4\lor \overline{x}_3x_4\) (\(x_3\neq x_4\)).}\label{fig:dnnf-a}
   \end{figure}
  
\begin{example}\label{ex:fullnnf}
   Figure~\ref{fig:dnnf-a} presents a DNNF for a boolean function, for which \textsf{FullNNF}
   described in~\cite{AGMES16} is not a PC encoding, although it implements domain
   consistency. The boolean input variables are $x_1, x_2, x_3, x_4, x_5$
   and the output is $\rho$.
  
   Encoding \textsf{FullNNF} uses main variables \(x_1, \dots, x_5\)
   and auxiliary variables which represent the inner gates of the
   DNNF\@. Note that partial assignment \(d_1\land d_2\) is contradictory, since we cannot
   have simultaneously $x_1=x_2$ and $x_1 \not= x_2$. However, unit
   propagation does not derive any literals except \(d_1\), \(d_2\)
   and \(\rho\), in particular it does not derive the contradiction.
   Let us look at the clauses in \textsf{FullNNF} which contain
   variables \(d_1\) and \(d_2\).
   \begin{itemize}
      \item Since \(d_1\) and \(d_2\) are \(\lor\) gates, they are in
         the following clauses of group~\ref{FN1}:
         \begin{align*}
            d_1&\to e_1\lor e_2 & d_2&\to e_3\lor e_4
         \end{align*}
      \item Since \(d_1\) and \(d_2\) are inputs to \(\land\) gates
         \(c_1\) and \(c_2\), they are in the following clauses of
         group~\ref{FN2}:
         \begin{align*}
            c_1&\to d_1 & c_2&\to d_2
         \end{align*}
      \item In addition, \(d_1\) and \(d_2\) are in the following
         clauses of group~\ref{FN3}:
         \begin{align*}
            d_1&\to c_1\lor c_4 & d_2&\to c_2\lor c_3\\
            e_1&\to d_1 & e_3&\to d_2\\
            e_2&\to d_1 & e_4&\to d_2
         \end{align*}
   \end{itemize}
None of the clauses listed above becomes unit or empty after satisfying
\(d_1\) and \(d_2\) and, hence, unit propagation cannot be
applied to the formula. In particular, the contradiction is not
derived which implies that \(\textsf{FullNNF}\) is not a URC encoding.
\end{example}

The vertices $d_1$ and $d_2$ in Example~\ref{ex:fullnnf} are DNFs, however, they can be any
mutually excluding DNNFs defined on the same set of variables. As pointed out
before, testing
satisfiability of a conjunction of two DNNFs is an NP-complete problem.
This means that there is no known construction of an encoding of polynomial
size that derives a contradiction
from $d_1 \wedge d_2$, if and only if this assignment is contradictory
with a partial assignment of the inputs.
Instead, we use the fact that \textsf{FullNNF} encoding does not
require that a variable representing a gate is assigned $1$, if the gate
evaluates to $1$, and extend the encoding so that the
partial assignment $d_1\wedge d_2$ always leads to a contradiction.
Our approach is a generalization of the idea used in the construction
of \textsf{CompletePath} encoding of MDDs which we describe next.

\textsf{CompletePath} encoding introduced in~\cite{AGMES16} is a PC encoding
of a multivalued decision diagram (MDD) generalizing an ordered binary
decision diagram (OBDD) to arbitrary finite domains.
Its construction uses the fact that an MDD is satisfied if and
only if there is a path from the root to a leaf labeled \(1\) which
consists only of activated edges. The encoding uses auxiliary variables
for the edges and vertices of the MDD and consists of the following
two parts.
\begin{itemize}
\item Clauses that describe local conditions implying that a model
of the encoding represents the set of vertices and edges of the
unique accepting path, if it exists.
\item Exactly-one constraints implying that at every level of
the MDD, exactly one vertex belongs to the model. These clauses are
implied by the previous group of clauses, however, they are needed
to achieve propagation completeness.
\end{itemize}
There are encodings of MDD whose models contain more vertices and edges
than only those that belong to the unique accepting path, however, they
are not propagation complete.
Our PC encoding takes the idea of using exactly-one constraints to achieve
propagation completeness of the encoding of the paths
used in \textsf{CompletePath} and generalizes it to the case of DNNFs. The
situation of DNNFs is different in that a minimal certificate of its satisfiability
does not have a form of a path, but a tree, because we must
make sure that every child of a satisfied \(\land\)-vertex is also
satisfied. We call this tree a minimal satisfying subtree and
introduce it in Section~\ref{sub:mst}. The encoding
of minimal satisfying subtrees that we describe later uses exactly-one
constraints on specific subsets of the vertices of the DNNF\@. In particular,
the vertices $d_1$ and $d_2$ in Figure~\ref{fig:dnnf-a} belong to one of such sets
and, hence, the encoding implies $\neg d_1 \vee \neg d_2$.

\section{Multivalued DNNFs and the Main Result}
\label{sub:mdnnf}

In this section, we shall describe a generalization of DNNFs to
multivalued domains which was earlier considered in~\cite{GS12}. We
shall call this generalization a multivalued DNNF (MDNNF) to
distinguish it from the usual DNNF which is used to represent a
boolean function. Later in this section, we formulate the main result
of our paper which is the fact that we can construct a PC encoding for
MDNNFs, the result holds for DNNFs on boolean variables as well.

Consider a set of variables \(\vek{x}=(x_1, \dots, x_n)\) where
the domain of $x_i$ denoted $\dom{x_i}$ is a non-empty finite set.
A constraint \(f(\vek{x})\) is a mapping
\begin{equation}\label{eq:constraint}
   f:\dom{x_1}\times\dots\times\dom{x_n}\to\{0, 1\},
\end{equation}
where $0$ and $1$ represent the truth values.
In order to describe a tractable representation of such constraints, we
introduce multivalued DNNF\@.
The difference between a DNNF and a multivalued DNNF is the
interpretation of the leaves, otherwise, they are the same.

Assume \(D\) is an acyclic directed graph as in a NNF
with vertices \(V\), root \(\rho\in V\), the set of edges \(E\), and the
set of leaves \(L\subseteq V\) and such that
the inner vertices are labeled by $\vee$ and $\wedge$.
We say that \(D\) is a multivalued NNF, if the leaves \(L\) are labeled
with unary constraints \(x_i=a\) instead of the literals.
We assume that each unary constraint \(x_i=a\)
is used as a label of at most one leaf.
Some of the unary constraints may be missing in $D$, however, we assume
that for each $i=1,\ldots,n$ at least one unary constraint on
the variable $x_i$ appears in $D$.

If all the constraint variables $\vek{x}$ are boolean and
we identify the unary constraint $x_i=1$ with the literal $x_i$
and the unary constraint $x_i=0$ with the literal $\neg x_i$, then
NNF is a special case of a multivalued NNF\@.

For simplicity, we assume that no leaf of \(D\) is
labeled with a constant unless \(D\) is a single vertex representing
a constant function. This can be done without loss of generality,
since one can always simplify \(D\) by
propagating constant values in the leaves, if they are not the root.
For the construction of our encodings, we assume that the root of $D$
is not a constant.

We say that a multivalued NNF \(D\)
\emph{represents} a constraint \(f(\vek{x})\) if for every assignment
\(\assign{a}\) of \(\vek{x}\) we have that \(D\) evaluates to
\(f(\assign{a})\) if each leaf is evaluated according to the
unary constraint in its label.

Following~\cite{GS12}, decomposability and smoothness are
required with respect to constraint variables \(x_1, \dots, x_n\).
For a vertex \(v\in V\) let us denote \(\var{v}\) the set of the variables
from $\vek{x}$ that appear in the leaves which can be reached from \(v\)
by a directed path.
More precisely, a variable \(x_i\in\vek{x}\) belongs to
\(\var{v}\) if there is a directed path from \(v\) to a leaf labeled with
a unary constraint \(x_i=a\) for a value \(a\in\dom{x_i}\). In particular,
by assumption we have that \(\var{\rho}=\vek{x}\).

\begin{definition}\label{def:decomp-smooth-multi}
   We define the following structural restrictions of
   multivalued NNFs.\mynobreakpar%
   \begin{itemize}
      \item We say that multivalued NNF \(D\) is \emph{decomposable}
      (multivalued DNNF, MDNNF), if
         for every vertex \(v=u_1\land\dots\land u_k\) the
         sets of variables \(\var{u_1}, \dots, \var{u_k}\) are
         pairwise disjoint.
      \item We say that MDNNF \(D\) is \emph{smooth} if for every vertex
         \(v=u_1\lor\dots\lor u_k\) we have
         \(\var{v}=\var{u_1}=\dots=\var{u_k}\).
   \end{itemize}
\end{definition}

Requiring decomposability and smoothness is essential to our construction.
Decomposability is a strong restriction. In particular, the satisfiability
test is polynomial for MDNNF, while it is NP-complete for a general
(multivalued) NNF\@. On the other hand,
for a given non-smooth MDNNF, we can construct an
equivalent smooth one in polynomial time by a simple generalization of
the algorithm for DNNFs described in~\cite{D01a}.

Smoothness is a property which simplifies interpretation of the models
of a multivalued DNNF as sets of satisfied leaves. Let us also point
out that when using an MDD to implement a constraint in CP in a way
supporting explanations of the conflicts, it is suggested to use
a complete MDD where each path tests all variables~\cite{GSS11}.
An MDD can be interpreted as a MDNNF with a special structure and then,
the MDD is complete if and only if the corresponding MDNNF is smooth.

An important step of our construction is a construction of
a \emph{cover by separators}.
A simple form of such a cover is used in order to achieve propagation
completeness in \textsf{CompletePath}~\cite{AGMES16} encoding for MDDs
where the cover by separators is formed by a partition of an acyclic graph
into levels (layers). This is sufficient, since the computation of
an MDD is represented by a single path. For MDNNF, the computation
has more complex structure, however, the idea of a partition
into levels generalized to a cover by separators can be used,
if for each variable $x_i$, we consider
the subgraph of $D$ induced on the scope of $x_i$ separately.

Assume an MDNNF $D$ and a variable \(x_i\). The scope of $x_i$ in $D$
is the set \(V_i\) of vertices $v$ satisfying $x_i \in \var{v}$.
Let \(D_i\) be the subgraph of \(D\) induced on \(V_i\).
Moreover, let us denote $L_i = L \cap V_i$ which is the
set of the leaves of $D_i$. By assumptions on $D$, we have
$1 \le |L_i| \le |\dom{x_i}|$.
Note that if \(v\in V_i\) is labeled with \(\land\), then by
decomposability exactly one of the successors of \(v\) belongs to \(V_i\)
as well. On the other hand, if \(v\in V_i\) is labeled with \(\lor\), then by
smoothness all the successors of \(v\) belong to \(V_i\).
If a vertex belongs to $V_i$, then all its predecessors in $V$ belong
to $V_i$ as well.

\begin{example}\label{ex:subgraphs}
   Consider DNNF \(D\) from Figure~\ref{fig:dnnf-a}. Then
   \(D_5\) is the induced subgraph on vertices \(V_5=\{\rho, a_1, a_2,
      x_5, \neg x_5\}\) and \(D_1\) is the induced subgraph on vertices
   \(V_1=\{\rho, a_1, a_2, b_1, b_2, c_1, c_2, c_3, c_4, d_1, d_2,
      e_1, e_2, e_3, e_4, x_1, \neg x_1\}\). Note also that \(D_2\)
   differs from \(D_1\) only by including the leaves \(x_2\) and \(\neg x_2\) instead of
   \(x_1\) and \(\neg x_1\).
\end{example}

\begin{definition} \label{def:separator-a} \leavevmode
   \begin{itemize}
      \item A subset of vertices $S \subseteq V_i$ is called a
         \emph{separator} in $D_i$, if every path in $D_i$ from the
         root to a leaf contains precisely one vertex from $S$.
      \item We say that $D$ is \emph{covered by separators}, if
         for each $i=1,\ldots,n$, there is a collection of separators
         $\mathcal{S}_i$ in $D_i$, such that the union of $S \in
         \mathcal{S}_i$ is $V_i$.
   \end{itemize}
\end{definition}

Not every MDNNF can be covered by separators. However, every MDNNF
can be efficiently transformed into an equivalent one which admits
a separator cover, see Section~\ref{sec:separators} for more detail.

The size of the encoding constructed using a separator cover depends
on the total size of all separators. Without loss of generality,
we can assume that this total size is polynomial by the following
argument. For each
$i \in \{1,\ldots,n\}$ and each vertex $v \in D_i$ choose
a separator $S \in \mathcal{S}_i$ containing $v$. The chosen
separators form a subset of the original cover which is itself
a separator cover.
For each $i=1,\ldots,n$, it consists of at most $|V_i|$ separators
in $D_i$ each of size at most $|V_i|$ and, hence, the cover has
polynomial total size. A better estimate will be formulated
in Proposition~\ref{prop:separators-complexity}.

A complete MDD or an MDD with no long edges as in~\cite{AGMES16}
can be considered as a special case of a smooth strictly
leveled MDNNF\@. An MDNNF is strictly leveled if for each $v \in V$,
all paths from $\rho$ to $v$ have the same length. For such an MDNNF, there is
a cover by separators in which for every $i \in \{1,\ldots,n\}$, every separator
in $D_i$ consists of all vertices at a given level and the leaves above this level.

\begin{example}\label{ex:separators}
   Consider DNNF \(D\) from Figure~\ref{fig:dnnf-a}. Recall from
   Example~\ref{ex:subgraphs} that \(D_1\) is the induced subgraph on
   vertices \(V_1=\{\rho, a_1, a_2, b_1, b_2, c_1, c_2, c_3, c_4, d_1,
      d_2, e_1, e_2, e_3, e_4, x_1, \neg x_1\}\). Since \(D\) is
   strictly leveled, we can form a separator cover of \(D_1\) with
   separators being the levels of \(D_1\). In particular
   \(\mathcal{S}_1\) is formed by the following separators
   \begin{align*}
      S_\rho&=\{\rho\} &
      S_a&=\{a_1, a_2\}\\
      S_b&=\{b_1, b_2\} &
      S_c&=\{c_1, c_2, c_3, c_4\}\\
      S_d&=\{d_1, d_2\} &
      S_e&=\{e_1, e_2, e_3, e_4\}\\
      S_x&=\{x_1, \neg x_1\} &&
   \end{align*}
   However, this is not the only way of defining a separator cover, we
   can replace \(S_d\) and \(S_e\) with the sets
   \begin{align*}
      S_1&=\{e_1, e_2, d_2\}\\
      S_2&=\{d_1, e_3, e_4\} \;.
   \end{align*}
\end{example}
\bigskip

Our goal is to construct a CNF encoding of constraint~\eqref{eq:constraint}
represented by an MDNNF\@.
We encode the vector $\vek{x}$ with the boolean variables
representing the \emph{direct encoding} (see for example~\cite{AGMES16,BKNW09})
of the elements of finite domains.
This encoding uses the domain variables $\litVar{x_i=a}$, where
$a \in \dom{x_i}$, which are assumed to represent the
truth value of the corresponding constraint $x_i=a$.
The vector of the domain variables related
to $x_i$ will be denoted $\dvar{x_i}$ and the vector of all the
domain variables will be denoted $\dvar{\vek{x}}$.
If we say that an encoding uses the direct encoding of variables,
then we assume that
it contains the \emph{direct encoding constraints} which are
the exactly-one constraint for the block of variables $\dom{x_i}$
for every $i=1,\ldots,n$. If an assignment of $\dvar{\vek{x}}$ satisfies
the direct encoding constraints, we say that the assignment is
\emph{direct encoding consistent (DE-consistent)}.

\begin{definition} \label{def:enc-general-constraint}
A CNF formula $\phi(\dvar{\vek{x}}, \vek{y})$ is an encoding
of the constraint~\eqref{eq:constraint} using the direct encoding of
the variables $\vek{x}$, if it is an encoding of the boolean function
$f_\mathrm{de}(\dvar{\vek{x}})$ defined as follows
\begin{itemize}
\item every model of $f_\mathrm{de}$ is DE-consistent,
\item a DE-consistent assignment of $\dvar{\vek{x}}$ is a model
of $f_\mathrm{de}$ if and only if it represents a model
of~\eqref{eq:constraint}.
\end{itemize}
\end{definition}

The encoding \textsf{FullNNF}~\cite{AGMES16} of a boolean function represented
by a DNNF can be easily adapted to an encoding of
a multivalued DNNF using the approach described in~\cite{GS12}.
Recall that \textsf{FullNNF} consists of the unit clause
\(\rho\) for the root of a DNNF and clauses of groups~\ref{FN1}
to~\ref{FN4} described in Table~\ref{tab:clauses:fullnnf}.
For an MDNNF \(D\), we use variables \(\litVar{x_i=a}\)
to represent the unary constraints in the
leaves and variables that represent the inner vertices (gates) of \(D\).
The encoding consists of the direct encoding constraints for the
variables $\dvar{\vek{x}}$, the
clauses \(\rho\),~\ref{FN1} to~\ref{FN3} from
Table~\ref{tab:clauses:fullnnf}, and instead of clauses~\ref{FN4}, we
include unit clauses \(\neg \litVar{x_i=a}\) for the unary constraints
\(x_i=a\) that are not associated with any leaf of \(D\).
We denote the last group of clauses as~\ref{N4} in Table~\ref{tab:clauses:nnf-a}.
Since this is a straightforward modification of the original
\textsf{FullNNF} encoding, we keep the name and refer to this modification
as \textsf{FullNNF} encoding of a multivalued DNNF\@.
Our encoding is based on this generalization of \textsf{FullNNF},
uses the same variables, and extends it by clauses which restrict its models.

We can now formulate our main result.
\begin{theorem} \label{thm:main:general}
A multivalued DNNF representing a constraint~\eqref{eq:constraint} can be
converted in polynomial time into a URC and a PC encoding of the
constraint using the direct encoding of the input variables $\vek{x}$.
\end{theorem}

Every PC encoding is also URC, however,
we include both a URC and a PC encoding in the theorem, since
we shall construct them as different encodings.
There is only a minor difference between the proof of URC and PC
property for the basic form of the encodings, however, the constructions
of the smaller variants of the encodings in Section~\ref{sec:amo-eo}
use quite different arguments.
The proof is constructive and the construction of the encodings from
an input MDNNF $D^0$ consists of the following three steps:

\begin{enumerate}[label={(\alph*)}]
\item \label{enum:constr-smooth} Construct a smooth MDNNF $\Ds$ equivalent to $D^0$,
\item \label{enum:constr-separators} extend $\Ds$ to an MDNNF $\Dc$ for which
a separator cover can be obtained and construct such a cover,
\item \label{enum:constr-encoding} construct a URC and a PC encoding
of $\Dc$ using the chosen separator cover.
\end{enumerate}

Step~\ref{enum:constr-smooth} can be carried out in polynomial time by
a simple generalization of the algorithm for DNNFs described
in~\cite{D01a}. Note that at most $n(s+d)$ additional vertices and edges are
added during the construction, where $s$ is the number of the vertices
of $D^0$ and $d$ is the maximum of $|\dom{x_i}|$ over $i=1,\ldots,n$.
Step~\ref{enum:constr-separators} can be
carried out in polynomial time by
Proposition~\ref{prop:separators-complexity}.

\begin{table}[t]
   \begin{center}
      \begin{tabular}{l l l}
         \toprule
         \multicolumn{1}{c}{group} & clause & condition\\
         \midrule
         \manuallabel[N1]{N1} 
         & \(v\to v_1\vee\dots\vee v_k\) &\(v=v_1\vee\dots\vee v_k\)\\
         \manuallabel[N2]{N2} 
         & \(v\to v_i\) &\(v=v_1\land\dots\land v_k\), \(i=1, \dots, k\)\\
         \manuallabel[N3]{N3} 
         & \(v\to p_1\vee\dots\vee p_k\) &\(v\) has incoming edges from
         \(p_1, \dots, p_k\)\\
         \manuallabel[N4']{N4} 
         & $\neg l$ & $l \in \dvar{x_i} \setminus L_i$, $i=1,\ldots,n$\\
         \manuallabel[N5]{N5} 
         & \(\amorep{S}\) & $S \in \mathcal{S}_i$, \(i=1, \dots, n\) \\
         \manuallabel[N6]{N6} 
         & \(\exonerep{S}\) & $S \in \mathcal{S}_i$, \(i=1, \dots, n\) \\
         \bottomrule
      \end{tabular}
   \end{center}
   \caption{Clauses used in the construction of our URC and PC
      encodings of a smooth MDNNF \(\Dc\) with a fixed collection $\mathcal{S}_i$
      of separators in each $\Dc_i$.}\label{tab:clauses:nnf-a}
\end{table}

In order to construct CNF encodings in step~\ref{enum:constr-encoding},
denote by $V$ the set of the vertices of $\Dc$. All the vertices in $V$ are
considered as boolean variables and the vector of these variables
will be denoted $\vek{v}$. The list of clauses used in our encodings
is presented in Table~\ref{tab:clauses:nnf-a}.
All the groups of clauses
in the table except~\ref{N4} are formulated using the variables as elements
of $\vek{v}$ for simplicity. However, the variables in the inner vertices
and the variables in the leaves are treated differently in the encoding.
The variables represented by inner vertices are used in the encoding themselves
and the vector of these variables will be denoted $\vek{y} \subseteq \vek{v}$.
The variables in the leaves are the variables
$\vek{v} \setminus \vek{y} = \bigcup_{i=1}^n L_i$
and they are identified with the corresponding variables from $\dvar{\vek{x}}$.
In particular, we have $L_i \subseteq \dvar{x_i}$.
Some of the variables from $\dvar{\vek{x}}$ may not appear as leaves of $\Dc$
and these are forced to $0$ by the clauses in group~\ref{N4}.

As already mentioned, our encodings are based on a generalization
of \textsf{FullNNF} encoding which
is formed by clauses in groups~\ref{N1}--\ref{N3},~\ref{N4} in
Table~\ref{tab:clauses:nnf-a}, the unit clause \(\rho\), and the
direct encoding constraints.
Let us consider the following extensions of this encoding
by further clauses from Table~\ref{tab:clauses:nnf-a}:
\begin{description}[labelindent=\parindent]
   \item[\(\psi_c(\dvar{\vek{x}}, \vek{y})\)] consists of
      \textsf{FullNNF} of $\Dc$ and the clauses in group~\ref{N5},
   \item[\(\psi_p(\dvar{\vek{x}}, \vek{y})\)] consists of
      \textsf{FullNNF} of $\Dc$ and the clauses in group~\ref{N6}.
\end{description}

The use of the at-most-one and exactly-one constraints is motivated
by~\textsf{CompletePath} encoding~\cite{AGMES16}
as explained in Section~\ref{sub:fullnnf}. The effect of these constraints
in an MDNNF is analyzed in Section~\ref{sub:mst} using the notion
of a minimal satisfying subtree of \(\Dc\).
In Section~\ref{sec:encoding}, we prove that $\psi_c$ is
a URC encoding of \(\Dc\), so it can verify consistency efficiently.
Similarly, we prove that $\psi_p$ is a PC encoding of \(\Dc\), so
it guarantees efficient propagation.

\begin{remark}\label{rem:dec-redundant}
The \textsf{FullNNF} encoding contains the direct encoding
constraints, since they are important to guarantee domain consistency
on the direct encoding variables. The encodings $\psi_c$ and $\psi_p$
are extensions of \textsf{FullNNF}, so they contain direct encoding
constraints as well. On the other hand, these constraints are
redundant in both encodings
in the sense that they are consequences of the remaining parts
of the encodings and they are not needed for the proof of the claimed
propagation strength.
In particular, \(\psi_p\) is a PC encoding of \(f_{\mathrm{de}}\)
even without explicitly added direct encoding constraints, so
these constraints do not increase propagation strength (they are
absorbed by the other clauses in the encoding in the sense of~\cite{BM12}).
Moreover, if the leaves $L_i$ of the DNNF for some
$1 \le i \le n$ contain all
possible values of $\dom{x_i}$ and $L_i$ is chosen as a separator,
then the direct encoding constraints for $\dvar{x_i}$ are precisely
the clauses~\ref{N6} for this separator.
The encoding \(\psi_c\) is a URC encoding of $f_{\mathrm{de}}$
even without explicitly added direct encoding constraints, however,
these constraints can increase
its propagation strength on the domain variables.
\end{remark}

Let us look at an example how including the cardinality constraints on
separators improves the strength of unit propagation.

\begin{example}
   Consider DNNF \(D\) from Figure~\ref{fig:dnnf-a}. In
   Example~\ref{ex:fullnnf}, we demonstrated that partial assignment
   \(d_1\land d_2\) is contradictory and yet unit propagation on
   \textsf{FullNNF} with this partial assignment does not derive
   contradiction. Consider a cover which uses separators
   \begin{align*}
      S_1&=\{e_1, e_2, d_2\}\\
      S_2&=\{d_1, e_3, e_4\}
   \end{align*}
   introduced in Example~\ref{ex:separators}.
   Let us show that \(\psi_c\land d_1\land d_2\vdash_1\bot\). Indeed,
   using \(\amorep{S_1}\) and \(d_2\), unit propagation derives \(\neg e_1\) and
   \(\neg e_2\). Using clause \(d_1\to e_1\lor e_2\) from
   group~\ref{N1}, unit
   propagation derives \(\neg d_1\) which together with \(d_1\)
   derives contradiction.

   The exactly-one constraints in \(\psi_p\) offer stronger derivation
   properties. The encoding $\psi_p$ contains \textsf{FullNNF} which semantically implies
at-least-one condition on every separator. It follows that
\(\psi_p\land \neg d_1\models d_2\), however, unit propagation
in \textsf{FullNNF} does not guarantee the corresponding derivations.
Let us show that
\(\psi_p\land\neg d_1\vdash_1 d_2\). Using clauses \(e_1\to d_1\)
   and \(e_2\to d_1\) from group~\ref{N3}, unit propagation derives
   \(\neg e_1\) and \(\neg e_2\). Then using \(\exonerep{S_1}\) which
   is part of \(\psi_p\), unit propagation derives \(d_2\). Note that
   the last step requires exactly-one constraints and
   the at-most-one constraint in \(\psi_c\) is not enough.
\end{example}

The encodings \(\psi_c\) and \(\psi_p\)
contain prime representations of at-most-one and exactly-one constraints,
respectively, for each separator.
This is sufficient to prove that the time complexity of the construction
is polynomial, however, the encoding contains the above cardinality constraints
used on large groups of variables and their prime representations have
size quadratic in the number of the variables.
The size of the encoding decreases, if linear size
encodings of at-most-one and exactly-one constraints are used.
In Section~\ref{sec:amo-eo} we present sufficient conditions under
which this leads to a URC or a PC encoding and in Section~\ref{sec:complexity}
we present estimates of the size of the resulting encodings.

A boolean variable \(x_i\) has the binary domain \(\dom{x_i}=\{0, 1\}\),
the variable \(\litVar{x_i=1}\) represents the positive literal \(x_i\),
and the variable \(\litVar{x_i=0}\) represents the negative literal \(\neg x_i\).
Let us demonstrate that the substitutions
$\litVar{x_i=1} \gets x_i$ and $\litVar{x_i=0} \gets \neg x_i$
applied to the encodings from Theorem~\ref{thm:main:general}
preserve their propagation strength.
Formally, the substitutions should be applied once the encoding is constructed,
however, one can use the literals instead of the domain variables already during the
construction, since the resulting encoding is the same.

\begin{corollary} \label{cor:main:boolean}
A DNNF representing a boolean function can be converted in polynomial time
into a URC and a PC encoding of the same function.
\end{corollary}

\begin{proof}
Assume, $f(\vek{x})$ is a boolean function.
Let $\theta(\dvar{\vek{x}},\vek{y})$ be a URC or PC encoding of
the boolean function $f_\mathrm{de}(\dvar{x})$
guaranteed by Theorem~\ref{thm:main:general}
or some of the formulas $\psi_c$ or $\psi_p$ with
the direct encoding constraints removed. As argued in
Remark~\ref{rem:dec-redundant},
the clauses of direct encoding constraints can be removed without
decreasing the required level of propagation strength.

Let us show that the formula $\theta''$ obtained from $\theta$ by the
substitutions $\litVar{x_i=1} \gets x_i$ and $\litVar{x_i=0} \gets \neg x_i$
is an encoding of $f$ which is URC or PC, respectively.
By definition of $f_\mathrm{de}$, this formula is an encoding of $f$.
In order to prove its propagation strength, it is more
convenient to rename the variable $x_i$ to $\litVar{x_i=1}$ in
$\theta''$ and we denote $\theta'$ the formula after this renaming.
Renaming a variable does not change the propagation strength and
$\theta'$ can be obtained from $\theta$ by a simpler substitution, namely
$\litVar{x_i=0} \gets \neg \litVar{x_i=1}$ for all $i=1,\ldots,n$.

Let $\Delta$ be the conjunction of the direct encoding constraints
$(\litVar{x_i=1} \vee \litVar{x_i=0}) \wedge
(\neg \litVar{x_i=1} \vee \neg \litVar{x_i=0})$
for $i=1,\ldots,n$.
Clearly, $\Delta \models \theta \equiv \theta'$
which implies that $\Delta \wedge \theta$ and
$\Delta \wedge \theta'$ are equivalent.

Assume, $\theta$ is a PC formula. Since
\(\theta\equiv\theta\land\Delta\) according to
Remark~\ref{rem:dec-redundant} and adding implicates to a PC formula does
not decrease its propagation strength, we have that
\(\theta\land\Delta\) is PC as well.
Let $\alpha \subseteq \lit{\var{\theta'}}$ and $l \in \lit{\var{\theta'}}$,
where $\var{\theta'}$ is the set of variables of $\theta'$ and
contains $\litVar{x_i=1}$ as a replacement of $x_i$. Moreover,
assume
$$
\theta' \wedge \alpha \models l \;.
$$
This implies
$$
\Delta \wedge \theta' \wedge \alpha \models l
$$
since $\Delta$ in this context is a definition of the variable
$\litVar{x_i=0}$ which is not used in $\theta'$, $\alpha$, and $l$.
Since $\Delta \wedge \theta$ and $\Delta \wedge \theta'$ are equivalent,
we obtain
$$
\Delta \wedge \theta \wedge \alpha \models l
$$
and since $\theta\land\Delta$ is PC, we have
$$
\Delta \wedge \theta \wedge \alpha \vdash_1 l
$$
or
$$
\Delta \wedge \theta \wedge \alpha \vdash_1 \bot \;.
$$

We can use \(\Delta\) in unit propagation to derive \(\litVar{x_i=0}\)
given \(\neg\litVar{x_i=1}\) and vice versa.
Similarly, we can derive \(\neg\litVar{x_i=0}\)
given \(\litVar{x_i=1}\) and vice versa.
It follows that every unit propagation
derivation in \(\Delta \land \theta\) can be simulated in
\(\Delta \land \theta'\) and also vice versa. In particular,
unit propagation derives the same set of
literals from the formulas $\Delta \wedge \theta \wedge \alpha$ and
$\Delta \wedge \theta' \wedge \alpha$.
Hence, we have
$$
\Delta \wedge \theta' \wedge \alpha \vdash_1 l
$$
or
$$
\Delta \wedge \theta' \wedge \alpha \vdash_1 \bot \;.
$$

It remains to show
$$
\theta' \wedge \alpha \vdash_1 l
$$
or
$$
\theta' \wedge \alpha \vdash_1 \bot \;.
$$
A literal on the variable $\litVar{x_i=0}$ for some $i$ can appear
in a derivation from $\Delta \wedge \theta' \wedge \alpha$ only
using some of the clauses $\litVar{x_i=1} \vee \litVar{x_i=0}$ and
$\neg \litVar{x_i=1} \vee \neg \litVar{x_i=0}$ and both of these clauses
are satisfied after this derivation step. It follows that the derived
literal cannot be used to derive any new literal in a later
step of the derivation. Hence, we can avoid the derivation
of this literal and using the clauses of $\Delta$ in the derivation
of $l$ or $\bot$.

If $\theta$ is a URC formula, we prove that $\theta'$ is a URC formula
in a similar way, however, we consider only derivation of $\bot$.
\end{proof}

\section{Minimal Satisfying Subtrees and Separator Covers}\label{sub:mst}

The set of the variables of \textsf{FullNNF} encoding of an MDNNF
contains all $\dvar{\vek{x}}$ variables, however, the variables not used
as leaves are forced to $0$ by clauses~\ref{N4}. Hence, we can disregard
them and identify models of \textsf{FullNNF} with sets of vertices
of an MDNNF corresponding to the variables with value $1$ in the model.

The clauses of \textsf{FullNNF} encoding guarantee that each of the models is a subset
of the set of vertices satisfied in an accepting computation of an MDNNF\@.
Moreover, the subgraph induced by this subset is sufficient
as a certificate of the fact that the MDNNF is satisfied by an assignment of
the input variables. This subgraph is not necessarily inclusion minimal.
In particular, a \(\lor\)-vertex in the subgraph can have two or more
successor vertices in the subgraph and one of them can be removed.
In order to obtain propagation completeness, we extend \textsf{FullNNF}
by additional clauses which restrict the set of models of the encoding
to inclusion minimal subgraphs that are certificates of satisfiability
called a minimal satisfying subtree introduced below.
Moreover, the resulting restrictions on the variables representing the
vertices of a minimal satisfying subtree can be enforced by unit propagation.

In this section, $\Ds$ is an arbitrary smooth MDNNF and $\Dc$ is
an arbitrary smooth MDNNF that is covered by separators. In both cases,
the MDNNF has the set of vertices \(V\), root
\(\rho\in V\), directed edges \(E\), and leaves \(L \subseteq V\)
and represents constraint~\eqref{eq:constraint} on
finite domain variables \(\vek{x}=(x_1, \dots, x_n)\). In particular,
the leaves are associated with unary constraints of the form $x_i = a$
for $a \in \dom{x_i}$. We denote $\Ds_i$ and $\Dc_i$ the subgraph of
$\Ds$ and $\Dc$, respectively, induced on the scope of the variable $x_i$.

\begin{definition} \label{def:min-sat-subtree}
A \emph{minimal satisfying subtree} \(T\) of \(\Ds\) is any subgraph of \(\Ds\)
which has the following properties:
\begin{enumerate}
   \item $T$ contains the root \(\rho\) of \(\Ds\).
   \item For every \(\land\)-vertex \(v\) in \(T\), all edges $(v,u)$ in $\Ds$
         are in \(T\).
   \item For every \(\lor\)-vertex \(v\) in \(T\), exactly one of the edges $(v,u)$
         in $\Ds$ is in \(T\).
      \item\label{enum:def:mst:4} For every vertex $v$ in $T$, $v \not=\rho$, there is an edge $(u,v)$ in $T$.
\end{enumerate}
\end{definition}

Condition~\ref{enum:def:mst:4} corresponds to clauses~\ref{FN3} used
in encodings of DNNFs that guarantee domain consistency. We could
require exactly one incoming edge to $v$, however, this is not
necessary, since this is a consequence of the definition and the
decomposability of \(D\), see Remark~\ref{rem:arborescence} below.

Let us note that a minimal satisfying subtree used as a certificate
of satisfiability can be defined also as a minimal satisfied sub-DNNF,
see e.g.~\cite{BCMS16}.
The following lemma states a basic property of minimal satisfying
subtrees that is important for our construction.

\begin{lemma} \label{lem:min-sat-subtree-1}
If $T$ is a minimal satisfying subtree of $\Ds$, then
for each $i=1,\ldots,n$, $T \cap \Ds_i$ is
a path from the root to a leaf.
\end{lemma}

\begin{proof}
Assume, $T$ is a minimal satisfying subtree.
Fix a variable $x_i$. By assumption on $\Ds$, $\rho \in \Ds_i$.
If $v \in T \cap \Ds_i$ and $v$ is a $\land$-vertex, then by decomposability,
exactly one of the successors of $v$ in $\Ds$ is in $\Ds_i$ and by
assumption on $T$, the edge from $v$ to this successor
is in $T \cap \Ds_i$. If $v \in T \cap \Ds_i$
and $v$ is a $\lor$-vertex, exactly one edge $(v,u)$ is in $T$, since
$T$ is a minimal satisfying subtree. By smoothness, $u \in T \cap \Ds_i$
and, since $\Ds_i$ is an induced subgraph, the edge $(v,u)$ is in $T \cap \Ds_i$.
It follows that there is exactly one maximal path $P_i$ in $T \cap \Ds_i$
starting in the root. By Definition~\ref{def:min-sat-subtree},
the leaves of \(T\) are also leaves of \(D\) and thus the path
ends in a leaf.

Every vertex $v \in T$ is reachable from the
root by a path in $T$. Moreover, for every vertex $u$ in this
path, we have $\var{u} \supseteq \var{v}$. It follows that every vertex
$v \in T \cap \Ds_i$ is reachable from the root by a path in $T \cap \Ds_i$.
Hence, $T \cap \Ds_i$ is equal to $P_i$ and the proposition follows.
\end{proof}

\begin{remark} \label{rem:arborescence}
A minimal satisfying subtree is an out-arborescence (a rooted directed subtree)
with root \(\rho\) and the leaves which are leaves of \(D\).
This can be seen as follows.
Lemma~\ref{lem:min-sat-subtree-1} implies that $T$ is the union of
paths $T \cap \Ds_i$ for $i=1,\ldots,n$. If paths $P_i$ and $P_j$
split at a vertex $v$, then it is a $\land$-vertex and by
decomposability, the parts of $P_i$ and $P_j$ after this vertex
contain vertices with sets of variables which are disjoint subsets of
$\var{v}$.
\end{remark}

One can also verify that
Lemma~\ref{lem:min-sat-subtree-1} can be reversed
in the following sense.
If \(\Ds\) is a smooth MDNNF and \(T\) is an arbitrary subgraph of \(\Ds\)
such that for every \(i=1, \dots, n\) the intersection
$T \cap \Ds_i$ is a directed path from the root to a leaf in $L_i$,
then \(T\) is a minimal satisfying subtree of \(\Ds\).
This can be considered as a basic idea behind the construction
of our encodings. Later in
Proposition~\ref{prop:encoding-only-vertices}, we formulate a similar
argument in a form more suitable for the proof of the properties
of the encodings.

The relationship between minimal satisfying subtrees and assignments of the
variables $\vek{x}$ is straightforward. By Lemma~\ref{lem:min-sat-subtree-1},
every minimal satisfying subtree contains for every $i=1,\ldots,n$ exactly
one leaf containing the variable $x_i$. It follows that for every
minimal satisfying subtree $T$ there is a unique
assignment $\vek{a}$ of $\vek{x}$, such that $a_i \in \dom{x_i}$ and
the unary constraints in the leaves of $T$ are satisfied. More precisely,
the leaves of $T$ are exactly the leaves of $\Ds$ associated with the
unary constraints satisfied by $\vek{a}$.
Moreover, we have the following.

\begin{proposition} \label{prop:min-sat-subtree-2}
If $\vek{a}$ is a total assignment of $\vek{x}$, then $f(\vek{a})=1$ if and
only if there is a minimal satisfying subtree whose leaves are exactly the
leaves of $\Ds$ associated with the unary constraints satisfied by $\vek{a}$.
\end{proposition}

\begin{proof}
Assume $f(\vek{a})=1$ and
consider the subgraph $G$ of $\Ds$ induced by the vertices evaluating to $1$
in the computation of $\Ds$ for $\vek{a}$. Then, a subset of $G$
satisfying the properties of the minimal satisfying subtree $T$ can be
obtained by traversing $G$ top down. When a $\vee$-vertex $v$ is reached,
an arbitrary edge $(v,u)$ in $G$ can be included into $T$.

For the opposite direction, assume a minimal satisfying subtree $T$
whose leaves are exactly the leaves of $\Ds$ associated with the unary
constraints satisfied by $\vek{a}$. When evaluating $\Ds$
on input $\vek{a}$, all vertices of $T$ are evaluated to $1$. Since
$T$ contains the root, we have $f(\vek{a})=1$.
\end{proof}

It follows that the minimal satisfying subtrees can serve as certificates
of acceptance by $\Ds$ for assignments of the variables $\vek{x}$.
The models of the encodings which we construct correspond to the characteristic
functions of the sets of vertices of minimal satisfying subtrees.
Such an encoding will be called
an \emph{encoding of minimal satisfying subtrees}.
It is sufficient to encode the set of vertices
of a minimal satisfying subtree, since the assignment accepted by the
tree depends only on its leaves.

A PC encoding of paths in an acyclic graph that can be partitioned into levels
can be constructed using exactly-one constraints on the levels.
This is used, for example, in \textsf{CompletePath}
encoding for MDDs~\cite{AGMES16}. It appears that it is possible to
generalize this approach to minimal satisfying subtrees in MDNNFs using
Lemma~\ref{lem:min-sat-subtree-1} by which a minimal
satisfying subtree is a union of paths, one for each variable $x_i$.
In order to encode the condition that $T \cap \Ds_i$ is a path
we use a separator cover instead of a partition into levels. This allows
to construct an encoding of an MDNNF without the requirement that
it is strictly leveled.

\begin{remark} \label{rem:not-urc-enc}
In order to restrict the models of \textsf{FullNNF} encoding~\cite{AGMES16}
to the minimal satisfying
subtrees, a separator cover is not needed, if we do not require propagation
completeness. The models of \textsf{FullNNF} satisfy all the properties of
a minimal satisfying subtree except that a $\lor$-vertex can have more than one
successor.
Let us discuss a smaller set of clauses that restricts the models
to minimal satisfying subtrees, but is not propagation complete.

The conditions describing a minimal satisfying subtree in $\Ds$ can be
expressed in a straightforward way as a CNF formula whose variables
are the vertices of $\Ds$. To be exact, restricting the variables of
the formula to the vertices and disregarding the edges of $\Ds$ is correct
only if $\Ds$ does not contain a \emph{transitive edge}. This is an
edge $(v,u)$, such that $\Ds$ contains also another path from $v$ to $u$.
One can prove that a transitive edge is redundant in an MDNNF,
so we can assume that $\Ds$ does not contain a transitive edge.

The formula mentioned above contains \textsf{FullNNF} and,
additionally, the clauses representing the constraint
$v\to\amo{u_1, \dots, u_k}$ for every vertex \(v=u_1\lor\dots\lor u_k\).
The satisfying assignments of this formula correspond precisely to
the characteristic functions of the sets of vertices of
minimal satisfying subtrees.

One can show that in a smooth MDNNF
a minimal satisfying subtree cannot contain two successors of
the same $\vee$-vertex. It follows that we can strengthen the formula
and use $\amo{u_1, \dots, u_k}$ without the assumption $v$
for every vertex $v=u_1\lor\dots\lor u_k$.
However, even this stronger formula is not suitable for our purposes, since
it is not URC in general.
For example, it is not URC for the DNNF
from Example~\ref{ex:fullnnf}, since the partial assignment $d_1 \wedge d_2$
still does not derive a contradiction by unit propagation.
The additional clauses are the at-most-one constraints on pairs of variables
\(\{a_1, a_2\}\), \(\{c_1, c_2\}\), \(\{c_3, c_4\}\), \(\{e_1, e_2\}\), \(\{e_3,
   e_4\}\), \(\{e_5, e_6\}\), \(\{e_7, e_8\}\).
All of these clauses remain binary under $d_1 \wedge d_2$
and do not contribute to unit propagation.
\end{remark}

Instead of the above local approach to enforce minimal satisfying
subtrees, we use a more global one using a separator cover.
As explained at the beginnig of this section,
we can identify an assignment of values to the variables of
\textsf{FullNNF} encoding with a set of
the vertices in \(V\) satisfied by the assignment.
We shall use the following property of \textsf{FullNNF} to prove the main
proposition of this section.

\begin{claim}\label{claim:phiD}
   Let \(M\subseteq V\) be a model of \textsf{FullNNF} encoding of
   \(\Ds\) and let \(T\) be the subgraph induced by \(M\).
   For every $i\in\{1,\ldots,n\}$ and every $v_0 \in T \cap \Ds_i$, there
   is a path in $T \cap \Ds_i$ from the root to a leaf containing $v_0$.
\end{claim}

\begin{proof}
If $v \in T \cap \Ds_i$ and $v \not= \rho$, then the clauses~\ref{FN3}
guarantee that there is a vertex $u \in T$, such that $(u,v) \in T$.
Moreover, $\var{u} \supseteq \var{v}$, so $u \in T \cap \Ds_i$.
By inductive use of this argument, we obtain a path in $T \cap \Ds_i$ from the
root to $v_0$.

If $v \in T \cap \Ds_i$ and $v$ is a $\land$-vertex, then exactly one of
its successors in $\Ds$ is in $\Ds_i$ and by clauses~\ref{FN2}, this
successor is in $T$.
If $v \in T \cap \Ds_i$ and $v$ is a $\lor$-vertex, then by clauses~\ref{FN1},
at least one of its successors is in $T$. By smoothness, this successor
is in $T \cap \Ds_i$.
By induction using these arguments, we obtain a path in $T \cap \Ds_i$
from $v_0$ to a leaf.
\end{proof}

Existence of a cover by separators has the following consequence
for a smooth MDNNF\@. Note that it implies that the DNNF does not contain
a transitive edge discussed in Remark~\ref{rem:not-urc-enc}.

\begin{claim} \label{lem:successors-incomparable}
If $u_1$ and $u_2$ are different successors of the same vertex
in $\Dc$, then there is no path from $u_1$ to $u_2$.
\end{claim}

\begin{proof}
Let $v$ be the common predecessor of $u_1$ and $u_2$. If $v$ is
a $\wedge$-vertex, the statement follows from decomposability.
If $v$ is a $\vee$-vertex, then by smoothness, there is an index $i$,
such that all the vertices $v$, $u_1$, and $u_2$ belong to $\Dc_i$.
Assume for a contradiction that there is a path from $u_1$ to $u_2$ in $\Dc$.
Clearly, this path belongs to $\Dc_i$.
Let $S$ be a separator in $\Dc_i$ containing $u_1$.
Let $P_1$ be a path from the root to a leaf in $\Dc_i$
going through $v$, $u_1$, and $u_2$.
Moreover, let $P_2$ be obtained from $P_1$ by skipping $u_1$
using the edge $(v,u_2)$.
Since $u_1$ is the only vertex in $S \cap P_1$, the intersection
$S \cap P_2$ is empty which contradicts the definition of a separator.
It follows that there is no path from $u_1$ to $u_2$ in $\Dc$.
\end{proof}

We close this section by proving that the formulas $\psi_c$ and $\psi_p$
are encodings of the minimal satisfying subtrees.

\begin{proposition} \label{prop:encoding-only-vertices}
Assume, $\Dc$ is smooth and covered by separators.
For every \(M \subseteq V\), the following are equivalent
\begin{enumerate}[label={(\alph*)}]
\item $M$ is a model of $\psi_c$,\label{enum:prop:eov:b}
\item $M$ is a model of $\psi_p$,\label{enum:prop:eov:a}
\item $M$ is the set of the vertices of a minimal satisfying subtree of \(\Dc\),\label{enum:prop:eov:c}
\end{enumerate}
\end{proposition}

\begin{proof}
\(\text{\ref{enum:prop:eov:b}} \implies \text{\ref{enum:prop:eov:c}}\).
Assume, $M$ is a model of $\psi_c$ and let $T$ be the subgraph induced by $M$.
Since $M$ is also a model of the \textsf{FullNNF} encoding of $\Dc$,
$T$ satisfies all the properties of a minimal satisfying subtree except that
a $\lor$-vertex can have more than one successor in $T$.
Let $v$ be any $\lor$-vertex of $T$ and let us prove using the assumption
on the separators that at most one of its successors is in $T$.

Assume for a contradiction that two different
successors $u_1, u_2$ of $v$ are in $M$. By smoothness,
we have $v, u_1, u_2 \in \Dc_i$ for some $i$.
By Claim~\ref{claim:phiD}, there is a path $P$ in $T \cap \Dc_i$
from the root to a leaf containing $u_1$.
By assumption, there is a separator $S$ in $\Dc_i$ containing $u_2$.
The separator $S$ contains a vertex from $P \subseteq T$, however,
this vertex is not $u_2$, since by Claim~\ref{lem:successors-incomparable},
$u_2$ is not in $P$.
This implies $|M \cap S| \ge 2$ which is a contradiction,
since $M$ is a model of $\psi_c$.
It follows that every $\lor$-vertex of $T$ has at most one successor
in $T$ and, hence, $T$ is a minimal satisfying subtree.

   \(\text{\ref{enum:prop:eov:c}} \implies \text{\ref{enum:prop:eov:a}}\).
If $T$ is a minimal satisfying subtree, then the set $M$ of its
vertices satisfies \textsf{FullNNF} encoding.
Moreover, for every $i = 1,\ldots,n$, $P_i = T \cap \Dc_i$
is a path by Lemma~\ref{lem:min-sat-subtree-1}.
If $S$ is a separator in $\Dc_i$, we have
$$
|M \cap S| = |P_i \cap S| = 1
$$
by the properties of the separators. It follows that $M$ is a model of $\psi_p$.

\(\text{\ref{enum:prop:eov:a}} \implies \text{\ref{enum:prop:eov:b}}\) is clear,
since $\psi_c$ is a subset of $\psi_p$.
\end{proof}


\section{Covering a MDNNF with Separators}
\label{sec:separators}

In this section, we describe step~\ref{enum:constr-separators}
of the proof of the main result. Namely,
we show that for every smooth MDNNF \(\Ds\) one
can construct in polynomial time an equivalent smooth MDNNF \(\Dc\) which
can be covered by separators and moreover, a separator cover of \(\Dc\)
can be constructed in polynomial time as well.

In order to guarantee a cover by separators, we include auxiliary vertices
into the MDNNF that subdivide some of its edges. The only purpose of
these vertices is to create additional auxiliary variables in the encoding.
We call them \emph{no-operation} vertices, they represent the identity, and,
formally, they are disjunctions with one argument.
We describe a procedure to add suitable no-operation vertices controlled by
a labeling $\Level:V \to \mathbb{N}$ which is \emph{edge consistent}.
By this we mean that $\Level(\rho)=0$ and
for every edge $(v,u)$ in $\Ds$, we have $\Level(v) < \Level(u)$.

If $\Ds$ is strictly leveled, then an edge consistent labeling $\Level_0$
can be obtained, where for every vertex $v$, $\Level_0(v)$ is the common
length of the paths from $\rho$ to $v$.
In this case, the separators in $\mathcal{S}_i$ can be obtained
as the sets of all vertices $v$ satisfying $\Level_0(v)=j$ and,
additionally, the leaves $v$ satisfying $\Level_0(v) < j$ for
some $j \ge 1$. Hence, in this case,
we get a cover by separators with no auxiliary vertices.
For example, as already pointed out, if a complete MDD is
considered as an MDNNF, it is a strictly leveled smooth MDNNF\@.

It is easy to construct an edge consistent labeling for an
arbitrary MDNNF, however,
it is unclear, how to choose a labeling that minimizes
the number of new auxiliary vertices. For this reason, we leave the labeling
unspecified, although we describe two simple examples later.

Assume a fixed labeling $\Level$. An extension $\Dc$ of $\Ds$ that
can be covered by separators is obtained as follows.
For each edge $(v,u)$ in $\Ds$, for which $\Level(v) + 2 \le \Level(u)$,
include a new vertex $u'$ representing no-operation, replace the
edge $(v,u)$ by edges $(v,u')$ and $(u',u)$, and assume $\var{u'} = \var{u}$.
The resulting MDNNF is $\Dc$ and it is clearly equivalent to $\Ds$.
Let us denote $\Dc_i$ the subgraphs of $\Dc$ defined similarly as $\Ds_i$ in $\Ds$.

\begin{proposition} \label{prop:separators-by-Level}
If $\Level:V \to \mathbb{N}$ is an edge consistent labeling of the vertices
of a smooth MDNNF $\Ds$ and $\Dc$ is the extension of $\Ds$ obtained using
$\Level$ as above, then $\Dc$ is smooth, has size at most $|V|+|E|$ and can
be covered by separators.
\end{proposition}

\begin{proof}
Clearly, $\Dc$ has at most $|V|+|E|$ vertices. If $u'$ is a no-operation vertex
subdividing $(v,u)$, we have $\var{u'} = \var{u}$. It follows that smoothness
is preserved.

Construct separators for $\Dc$ as follows.
For every $i=1,\ldots,n$, let $d_i$ be the maximum of $\Level(v)$
for $v \in \Ds_i$. For every $i=1,\ldots,n$ and for each $j$, $0 \le j \le d_i$,
let $S_{i,j}$ be the union of the following three sets:
the set of the vertices $v \in \Ds_i$ satisfying $\Level(v)=j$,
the set of the leaves $v \in L_i$ satisfying $\Level(v) < j$,
and the set of all no-operation vertices used to subdivide some of the edges
$(v, u)$ in $\Ds_i$ satisfying $\Level(v) < j < \Level(u)$.
For each $i=1,\ldots,n$, let $\mathcal{S}_i$ be the collection
of the sets $S_{i,j}$ for $0 \le j \le d_i$.

Fix $i \in \{1,\ldots,n\}$ and $j$, $0 \le j \le d_i$.
Every path from $\rho$ to a leaf in $\Dc_i$ contains
either a single vertex $v \in \Ds$, such that $\Level(v)=j$,
a single leaf $v$ satisfying $\Level(v) < j$,
or a single pair of vertices $v,u \in \Ds_i$, such that $(v,u)$
is an edge of $\Ds_i$ and $\Level(v) < j < \Level(u)$.
Each of these cases implies a vertex of the path that belongs to $S_{i,j}$.
Since the cases exclude each other, $S_{i,j}$ is a separator in $\Dc_i$.
Moreover, every vertex of $\Dc_i$ is contained in $S_{i,j}$ for some $j$.
\end{proof}

Note that the construction used in the proof
of Proposition~\ref{prop:separators-by-Level}
yields $L_i \in \mathcal{S}_i$, since $S_{i,j}=L_i$ if $j=d_i$.
Let us estimate the size of the constructed separator cover
and the complexity of the construction.
By height of a MDNNF we mean the number of edges
on a longest path from the root to a leaf.

By the width of an MDNNF corresponding to a chosen labeling $\Level$,
we mean the maximum over $i=1,\ldots,n$ of the maximum size of
a separator in $\Ds_i$ constructed as in the
proof of Proposition~\ref{prop:separators-by-Level}. We call this maximum
the width, since it is also the size of a largest cut in any of the graphs
$\Ds_i$ at a given level defined as the set of the vertices at this level,
edges crossing this level, and leaves above this level.

\begin{proposition}\label{prop:separators-complexity}
Assume \(\Ds\) is a smooth MDNNF which represents a constraint on
\(n\) variables \(\vek{x}\) and has height \(h\). Let
\(\Level:V\to\mathbb{N}\) be an edge consistent labeling of the
vertices such that \(\Level(v) \le h\) for every $v\in V$.
Let $w$ be the width of $\Ds$ corresponding to the labeling $\Level$.
Then there is a separator cover \(\mathcal{S}=\bigcup_{i=1}^n\mathcal{S}_i\)
such that
\begin{itemize}
\item $L_i \in \mathcal{S}_i$ for every $i=1,\ldots,n$,
\item the total size of all
different separators is \(t=\sum_{S\in \mathcal{S}}|S|\leq n w h + 1\)
\item the cover can be constructed in time \(O(t)\).
\end{itemize}
\end{proposition}

\begin{proof}
Let \(\mathcal{S}=\bigcup_{i=1}^n\mathcal{S}_i\) be the separator
cover obtained by the construction in the proof of
Proposition~\ref{prop:separators-by-Level}.
One can verify that the first requirement is satisfied by inspecting the
proof of Proposition~\ref{prop:separators-by-Level}.
For each \(i=1, \dots, n\), the separator cover \(\mathcal{S}_i\) consists
of at most \(h\) nontrivial separators, each of which has size at most \(w\).
One trivial separator consisting only of the root vertex \(\rho\) is part of
\(\mathcal{S}\) and thus \(t\leq n w h + 1\).
The complexity of the construction \(O(t)\) is clear from the
construction in the proof of Proposition~\ref{prop:separators-by-Level}.
\end{proof}

Each of the parameters \(w\) and $h$ can be bounded from above by the
total number of vertices and edges in the input MDNNF which implies
that the construction can be done in time polynomial in the input size.
However, the parameters \(w\) and $h$ give a better insight into the size of the
separator cover constructed in this way.

Let us present two examples of an edge consistent labeling bounded
from above by the height of $\Ds$.
A simple example is the
function $\Level_1(v)$ defined for every vertex $v$ as the maximum
length of a path from the root to $v$. Since \(\Ds\) is acyclic, the
values of \(\Level_1(v)\) can be computed in linear time for every
vertex \(v\).
Another function that can be
used is $\Level_2(v)$ that is equal to $\Level_1(v)$ for the leaves,
however, for an inner vertex $v$, it is defined as
$$
\Level_2(v) = \min_{(v,u) \in \Ds} \Level_2(u) - 1 \ .
$$
Note that $\Level_2(v)$ is the largest possible label of $v$ if the
labels of the successors of $v$ are given and the labeling is
edge consistent.
The values of \(\Level_2(v)\) can be computed in linear time
by following the vertices in the reversed topological order of \(\Ds\).

\begin{remark*}
Using $\Level_2(v)$, the levels assigned to the vertices of a path from the
root to a leaf tend to concentrate on the values closer to the level of
the leaf. This implies that the edges close to the leaves have higher
chance to satisfy $\Level(v) + 1 = \Level(u)$ for \(\Level=\Level_2\)
than for \(\Level=\Level_1\). These edges then do not need to be
subdivided by auxiliary vertices.
\end{remark*}


\section{PC and URC Encodings of MDNNF}
\label{sec:encoding}

We are ready to show the main properties of encodings \(\psi_c\) and
\(\psi_p\) introduced in Section~\ref{sub:mdnnf} and used
in step~\ref{enum:constr-encoding} of the proof of the main result.
As in the previous sections, \(\Dc\) is a smooth MDNNF that can be
covered by separators. For each $i=1,\ldots,n$, consider
a fixed collection of separators $\mathcal{S}_i$ in $\Dc_i$.

Models of
our encodings exactly represent the set of minimal satisfying subtrees
of \(\Dc\). Since \(\Dc\) is smooth, the leaves of a minimal satisfying
subtree specify a full assignment of the variables in \(\vek{x}\)
by a vector of values from
\(\dom{x_1}\times\dots\times\dom{x_n}\). If a unary constraint
\(x_i=a\) is not associated with any leaf of \(\Dc\), then \(a\) cannot
be assigned as a value of \(x_i\) by any minimal satisfying subtree of
\(T\). The encoding thus forbids these assignments explicitly by
clauses of group~\ref{N4}. This ensures that every variable gets
exactly one value from its domain in every model of each of the encodings.
Equivalently, every model of the encodings specifies a DE-consistent
assignment of values in \dvar{\vek{x}}.

\begin{proposition} \label{prop:enc-correctness}
Each of the formulas \(\psi_c\) and \(\psi_p\) is an encoding of
the constraint \(f(\vek{x})\) represented by \(\Dc\).
\end{proposition}

\begin{proof}
By Proposition~\ref{prop:encoding-only-vertices},
\(\psi_c\) and \(\psi_p\) have the same set of models,
   in other words, they represent the same boolean function of
   variables \(\dvar{\vek{x}}\) and \(\vek{y}\). It is thus enough to
   show that \(\psi_p\) is a CNF encoding of \(\Dc\).
   By Definition~\ref{def:enc-general-constraint}, this means to
   show that \(\psi_p\) is a CNF encoding of the boolean function
   \(f_\mathrm{de}(\dvar{\vek{x}})\).

Assume, \(\assign{a}\) is an assignment of $\dvar{x}$ and $\vek{a}'$
is an assignment of $\vek{x}$ encoded by $\vek{a}$.
If $\vek{a}$ is a model of \(f_\mathrm{de}\), then $\vek{a}'$ is
accepted by $\Dc$ and by Proposition~\ref{prop:min-sat-subtree-2},
there is a minimal satisfying subtree \(T\) which is consistent
with \(\assign{a}'\) in that the leaves of \(T\) are labeled with
exactly the unary constraints satisfied by \(\assign{a}'\). This
means that the leaves of $T$ are precisely the leaves of $\Dc$
satisfied by $\vek{a}$.
By Proposition~\ref{prop:encoding-only-vertices}, the set $M$ of the
vertices of \(T\) is a model of $\psi_p$. Since this model
agrees with $\vek{a}$ on the variables $\dvar{x}$,
\(\psi_p(\assign{a}, \vek{y})\) is satisfiable.

   Assume on the other hand a model $(\vek{a},\vek{b})$ of \(\psi_p\)
   where \(\assign{a}\) assigns values to \(\dvar{\vek{x}}\) and
   \(\assign{b}\) assigns values to \(\vek{y}\).
By Proposition~\ref{prop:encoding-only-vertices}, variables with value $1$
in $(\vek{a},\vek{b})$ represent the set of vertices
of a minimal satisfying subtree \(T\) of \(\Dc\) and the leaves of $T$
are consistent with \(\assign{a}\). In particular \(\assign{a}\) is
DE-consistent and the existence of \(T\) certifies that \(\Dc\) with input
encoded by \(\assign{a}\) evaluates to true. Assignment \(\assign{a}\) is
thus a model of $f_\mathrm{de}$.
\end{proof}

In the rest of this section, we prove the claimed propagation strength of
$\psi_c$ and $\psi_p$.
Since the propagation is considered on all variables, it is a property
of the formula and we may disregard the encoded constraint. Due to this,
it is simpler not to distinguish the main variables $\dvar{\vek{x}}$
and the auxiliary variables $\vek{y}$, and we consider
the encodings as formulas $\psi_c(\vek{v})$ and $\psi_p(\vek{v})$
whose variables are the vertices of $\Dc$.
This is possible, since the only clauses which cannot be treated in this
way are the clauses in group~\ref{N4}. These clauses are unit clauses
on the $\dvar{\vek{x}}$ variables not used in $\Dc$, so they do not share
variables with the remaining parts of $\psi_c$ and $\psi_p$.
It follows that we may
ignore them for the proof of the propagation strength.

Let us prove that $\psi_c$ and $\psi_p$ are URC and
complete for deriving negative literals.

\begin{lemma}
   \label{lem:dnnf-notbot-sat} 
Let $\psi$ be any of the formulas $\psi_c$ and $\psi_p$.
Let \(\alpha\subseteq\lit{V}\) be a partial assignment, such that
\(\psi\land\alpha\not\vdash_1\bot\).
Then $\psi\land\alpha$ is satisfiable.
If, additionally, \(v_0\in V\) is a vertex such that
\(\psi\land\alpha\not\vdash_1\neg v_0\), then
\(\psi\land\alpha\land v_0\) is satisfiable.
\end{lemma}
\begin{proof}
It is sufficient to prove the second statement, since the
first statement follows from the second one using $v_0=\rho$.
Let \(V'=\{v\in V\mid \psi\land\alpha\not\vdash_1\neg v\}\) and
let \(D'\) be the subgraph of \(\Dc\) induced by the set \(V'\).
Since unit clause \(\rho\) is contained in \(\psi\) and
\(\psi\land\alpha\not\vdash_1\bot\), we have \(\rho\in V'\). By
assumption, \(v_0\in V'\).

   Let us show
that there is a path \(P\) from the root
\(\rho\) to \(v_0\) in \(D'\) by constructing \(P\) backwards
starting in \(v_0\). At the beginning, \(P\) is initialized as \(\{v_0\}\)
and then we repeat the following until we get to the root. Let
\(v\) be the first vertex of \(P\) and assume it is not the
root. Since \(v\in V'\), it has a predecessor \(u\)
which is in \(V'\) and we prepend \(u\) to \(P\).
The vertex $u$ exists, since otherwise \(\neg v\)
   would be derived by unit propagation using the clause of
   group~\ref{N3} corresponding to \(v\).

Let us construct a minimal satisfying subtree \(T\) of \(D'\) containing
$P$ by starting at the root $\rho$ and successively extending $T$ downwards.
The set of the vertices of $T$ is initialized as $\{\rho\}$ and
while there is a leaf \(v\) of \(T\) which is not a leaf of \(\Dc\),
we extend \(T\) as follows.
   \begin{itemize}
\item Assume \(v=u_1\lor\dots\lor u_k\). Since \(v\in T\),
we have that \(v\in V'\). If \(v\in P\) and \(v\neq v_0\),
then \(v\) has a successor \(u_j\) which belongs to \(P\) and
we add \(u_j\) and $(v, u_j)$ to \(T\). If \(v\not\in P\) or \(v=v_0\),
then there is a successor \(u_j\) of \(v\) which belongs to \(V'\)
and we add \(u_j\) and $(v, u_j)$ to \(T\). The vertex $u_j$ exists, since otherwise,
the clause in group~\ref{N1} corresponding to \(v\) derives \(\neg v\)
by unit propagation.
\item Assume \(v=u_1\land\dots\land u_k\). Since \(v\in T\),
it belongs to \(V'\) and also all its successors \(u_j\) belong to \(V'\).
Otherwise the clause \(v\to u_j\) in group~\ref{N2} derives \(\neg v\)
by unit propagation. We add all vertices \(u_1, \dots, u_k\) and
the corresponding edges to $T$.
   \end{itemize}
   It follows from the construction that \(T\) is a minimal satisfying
   subtree of \(\Dc\) in which all vertices belong to \(V'\).
Moreover, by construction, $T$ contains $P$ and, hence, also $v_0$.

By Proposition~\ref{prop:encoding-only-vertices} we have that the set of
vertices in \(T\) specifies a model of \(\psi\), let us denote it
\(\assign{a}\). Since \(v_0\) is in \(T\) by construction,
\(\assign{a}(v_0)=1\).
It remains to show that \(\assign{a}\) is consistent
with \(\alpha\).

If \(\neg v\in\alpha\) for some vertex \(v\in V\), then
\(v\not\in V'\) and thus \(v\) is not in \(T\). Assume a
positive literal \(v\in\alpha\) for a vertex \(v\in V\).
If $v$ is the root, then it belongs to $T$ by construction.
Otherwise, consider
an index \(i\in\{1, \dots, n\}\), such that \(v\in \Dc_i\).
There is a separator $S \in \mathcal{S}_i$, such that \(v \in S\).
Since the clauses in group~\ref{N5} or group~\ref{N6} are satisfied,
every vertex \(u\in S \setminus \{v\}\) satisfies
\(\psi\land\alpha\vdash_1 \neg u\) and thus \(u\not\in V'\).
It follows that \(T\) contains \(v\), since by
Lemma~\ref{lem:min-sat-subtree-1},
the intersection $T \cap \Dc_i$ is
a path from the root to a leaf and, hence, has
a nonempty intersection with $S$. Thus \(\assign{a}\) satisfies all
literals from \(\alpha\).
It follows that \(\assign{a}\) is a model of
\(\psi\land\alpha\land v_0\) and it is thus satisfiable.
\end{proof}

The following is specific to \(\psi_p\) encoding.

\begin{lemma}
   \label{lem:dnnf-psi-p} 
   Encoding \(\psi_p\) is propagation complete.
\end{lemma}
\begin{proof}
   Let \(\alpha\subseteq\lit{V}\) be a partial
   assignment satisfying \(\psi_p\land\alpha\not\vdash_1 \bot\). Let
   \(l\) be a literal for which \(\psi_p\land\alpha\not\vdash_1 l\).
   We shall show that
   \(\psi_p\land\alpha\land\neg l\) is satisfiable.
Let \(v\) be the vertex in literal \(l\), so \(l=v\) or \(l=\neg v\).
If \(l=\neg v\), then \(\psi_p\land\alpha\land v\) is satisfiable
by application of Lemma~\ref{lem:dnnf-notbot-sat} with \(v_0=v\).
For the rest of the proof, assume $l=v$.

   Since \(\psi_p\land\alpha\not\vdash_1\bot\),
\(\psi_p\land\alpha\) is satisfiable by Lemma~\ref{lem:dnnf-notbot-sat}.
In particular,
   $\psi_p\wedge\alpha$ cannot simultaneously imply
   $v$ and $\neg v$. Thus if
   \(\psi_p\land\alpha\vdash_1\neg v\), then
   \(\psi_p\land\alpha\not\models v\) and
   \(\psi_p\land\alpha\land\neg v\) is satisfiable as required.
For the rest of the proof, assume \(\psi_p\land\alpha\not\vdash_1\neg v\)
in addition to the assumption $\psi_p\land\alpha\not\vdash_1 v$.

   Let \(i\in\{1, \dots, n\}\) be an index, such that \(v\in V_i\)
   and let $S \in \mathcal{S}_i$ be a separator containing $v$.
   There must be another vertex \(v'\in S\) for which neither
   \(\psi_p\land\alpha\not\vdash_1 v'\), nor
   \(\psi_p\land\alpha\not\vdash_1\neg v'\),
   since otherwise the clauses in group~\ref{N6} derive $v$ or $\neg v$.
   By Lemma~\ref{lem:dnnf-notbot-sat} applied with \(v_0=v'\),
   we get that formula \(\psi_p\land\alpha\land v'\)
   is satisfiable. Since \(v\) and \(v'\) are both in the same
   separator \(S\), clauses~\ref{N6} imply that any model of \(\psi_p\land\alpha\land
      v'\) falsifies \(v\) and thus \(\psi_p\land\alpha\land\neg v\)
   is satisfiable.

In all cases we obtained that \(\psi_p\land\alpha\land\neg l\) is
satisfiable as required.
\end{proof}

Let us summarize the results of this section as follows.

\begin{theorem}
\label{thm:dnnf} 
Formula \(\psi_c(\dvar{\vek{x}}, \vek{y})\) is a URC encoding
and formula \(\psi_p(\dvar{\vek{x}}, \vek{y})\) is a PC encoding of
constraint~\eqref{eq:constraint}. Both of them can be constructed
in polynomial time.
\end{theorem}

\begin{proof}
By Proposition~\ref{prop:enc-correctness}, the formulas
$\psi_c$ and $\psi_p$ are encodings of constraint~\eqref{eq:constraint}.
   Encoding \(\psi_c\) is URC by Lemma~\ref{lem:dnnf-notbot-sat} and
   encoding \(\psi_p\) is PC by Lemma~\ref{lem:dnnf-psi-p}.

   The separator cover for \(\Dc\) can be obtained in polynomial time by
   Proposition~\ref{prop:separators-complexity}. The construction of
   both encodings can clearly be carried out in time which is polynomial in the
   size of these encodings.
In order to estimate this size, we assume that the separator cover is
constructed as in Proposition~\ref{prop:separators-by-Level} using
a labeling $\Level$ bounded by the height of $\Dc$.
   
   Denote
   \(d=\max_{i=1}|\dom{x_i}|\) the maximum size of a domain, \(s\)
   the number of vertices of \(\Dc\), \(e\) the number of
   edges of \(\Dc\) and $h$ the height of $\Dc$.
Both encodings have at most \(nd+s\) variables,
   since \(|\dvar{\vek{x}}|\leq nd\) and \(|\vek{y}|\leq s\).
   
   The total number of clauses in groups~\ref{N1} to~\ref{N3} is \(O(e)\).
   The number of clauses in group~\ref{N4} is at most \(nd\) and the
   number of clauses in group~\ref{N5} is at most \(s^2\), because
   every variable is contained in at most \(s-1\) negative binary clauses.
   Clauses of group~\ref{N6} include an additional clause for each
   separator \(S\) consisting of all its variables, the number of
   these clauses is thus bounded by \(|\mathcal{S}|\leq nh + 1 \le ns + 1\).

   It follows that \(\psi_c\) and \(\psi_p\) consist of
   \(O(e+nd+s^2+ns)\) clauses which is polynomial in the size of
   \(\Dc\).
The direct encoding constraints are not included in this estimate, since
they are not needed for the proof of the claimed properties of the
encodings. On the other hand, they have polynomial size, so the size of
the encoding together with these constraints is also polynomial.
\end{proof}

For simplicity of proving the required propagation strength, the
encodings used in Theorem~\ref{thm:dnnf} contain
prime representations of the at-most-one or the exactly-one constraint
on each separator.
Due to this, the upper bound on the size of the encodings
is quadratic in the number of the vertices of $\Dc$.
In Section~\ref{sec:complexity}, we prove that this is not necessary
if we use suitably chosen encodings of linear size of the two
cardinality constraints.

\section{Embedding Basic Cardinality Constraints in an Encoding}
\label{sec:amo-eo}

CNF encodings of cardinality constraints frequently use auxiliary variables
since this allows to reduce the size of the encoding.
The main topic of this section is to investigate the consequences
of using auxiliary variables in the at-most-one and the exactly-one
constraints for the propagation strength of a larger encoding
containing the constraint in question as its part.
Sufficient conditions for using auxiliary variables for this purpose
are presented in sections~\ref{sub:amo} and~\ref{ssec:exactly-one}
and an estimate of the size of the resulting encodings is
in Section~\ref{sec:complexity}.

\subsection{Preliminary Considerations}

Let us start with a more general question.
Consider an encoding $\varphi(\vek{x}, \vek{x}', \vek{y}) \land \theta(\vek{x})$
of a constraint on the variables $\vek{x} \cup \vek{x}'$ and let
$\theta'(\vek{x}, \vek{z})$ be an encoding of the subformula $\theta(\vek{x})$
using new auxiliary variables $\vek{z}$. Then replacing
$\theta$ by $\theta'$ yields an encoding
$\varphi(\vek{x}, \vek{x}', \vek{y}) \land \theta'(\vek{x}, \vek{z})$
of the same constraint on the variables $\vek{x}\cup\vek{x}'$, because
\begin{equation} \label{eq:general-embedding}
(\exists \vek{y})[\varphi(\vek{x}, \vek{x}', \vek{y})\land\theta(\vek{x})]
\equiv
(\exists \vek{y})(\exists \vek{z})[\varphi(\vek{x}, \vek{x}',
\vek{y})\land\theta'(\vek{x}, \vek{z})]\,.
\end{equation}
A general question is what can be said about
propagation strength of the right-hand side using
assumptions on the propagation strength of the left-hand side
and the properties of $\theta'(\vek{x}, \vek{z})$.

There is a significant
difference between replacing a subformula of an encoding if the
required propagation strength is implementing domain consistency
and if it is propagation completeness.
In the case of domain consistency, the natural necessary conditions
are also sufficient by Proposition~\ref{prop:DC-encoding}.
For technical reasons, the proposition is formulated in terms
of propagation completeness on the main variables of the encoding
which is equivalent to implementing domain consistency.
On the other hand, we demonstrate by an example that
a similar statement is not true for propagation completeness on
all the variables of an encoding.

\begin{proposition} \label{prop:DC-encoding}
Assume that \(\theta'(\vek{x}, \vek{z})\) is an encoding of \(\theta(\vek{x})\)
which is PC on variables \(\vek{x}\).
Assume \(\varphi(\vek{x}, \vek{x}', \vek{y})\land\theta(\vek{x})\)
is an encoding of a constraint on the variables \(\vek{x} \cup \vek{x}'\)
that is PC on these variables. Then
\(\varphi(\vek{x}, \vek{x}', \vek{y})\land\theta'(\vek{x}, \vek{z})\) is
an encoding of the same constraint on the variables \(\vek{x}\cup\vek{x}'\)
that is also PC on these variables.
\end{proposition}

\begin{proof}
Using~\eqref{eq:general-embedding}, we obtain that
$\phi \wedge \theta$ and $\phi \wedge \theta'$
are encodings of the same constraint on the variables $\vek{x}\cup\vek{x}'$.
   Let us show that \(\varphi(\vek{x}, \vek{x}',
      \vek{y})\land\theta'(\vek{x}, \vek{z})\) is PC on variables
   \(\vek{x}\cup\vek{x}'\).
For the rest of the proof, let \(\alpha\subseteq\lit{\vek{x}\cup\vek{x}'}\)
   be a partial assignment and \(l\in\lit{\vek{x}\cup\vek{x}'}\) be a
   literal such that
   \(\varphi(\vek{x}, \vek{x}', \vek{y})\land\theta'(\vek{x},
      \vek{z})\land\alpha\models l\).
Our goal is to show that $l$ or $\bot$ can be derived by unit propagation.

By~\eqref{eq:general-embedding}, we have
   \(\varphi(\vek{x}, \vek{x}',
      \vek{y})\land\theta(\vek{x})\land\alpha\models l\). Since
   \(\varphi\land\theta\) is PC on variables \(\vek{x}\cup\vek{x}'\),
   we have \(\varphi(\vek{x}, \vek{x}',
      \vek{y})\land\theta(\vek{x})\land\alpha\vdash_1 \bot\) or
   \(\varphi(\vek{x}, \vek{x}',
      \vek{y})\land\theta(\vek{x})\land\alpha\vdash_1 l\).
   Any of these two unit propagation derivations can be
   simulated by unit propagation in \(\varphi\land\theta'\).
   This is can be seen as follows. If a clause of
\(\theta\) is used to derive a literal \(e\in\lit{\vek{x}}\)
from assumptions \(\beta\subseteq\lit{\vek{x}}\), then
\(\theta\land\beta\vdash_1 e\). This implies \(\theta'\land\beta\vdash_1 e\)
or $\theta'\land\beta\vdash_1 \bot$,
because \(\theta'\) is PC on variables \(\vek{x}\).
In particular, using a clause of \(\theta\) in the original
derivation can be replaced with a sequence of
propagation steps in \(\theta'\) which either derive the same
literal or a contradition.
We thus get that $l$ or $\bot$ can be derived from
\(\varphi(\vek{x}, \vek{x}', \vek{y})\land\theta'(\vek{x}, \vek{z})\land\alpha\)
by unit propagation as required.
\end{proof}

Let us now consider propagation completeness.
The assumption that the encodings $\phi \wedge \theta$ and $\theta'$
are PC is not sufficient to guarantee that $\phi \wedge \theta'$ is even URC\@.
Let us demonstrate this by the following example, where $\theta(\vek{x})$
represents the at-most-$2$ constraint. Consider the formula
$$
\phi(\vek{x}) =
(x_1 \vee x_2 \vee x_3)
(x_1 \vee x_2 \vee x_4)
(x_1 \vee x_3 \vee x_4)
(x_2 \vee x_3 \vee x_4)
$$
for the at-least-$2$ constraint on the variables $(x_1, x_2, x_3, x_4)$
and the formula
$$
\theta(\vek{x}) =
(\neg x_1 \vee \neg x_2 \vee \neg x_3)
(\neg x_1 \vee \neg x_2 \vee \neg x_4)
(\neg x_1 \vee \neg x_3 \vee \neg x_4)
(\neg x_2 \vee \neg x_3 \vee \neg x_4)
$$
for the at-most-$2$ constraint on these variables.
Let $\theta'(\vek{x}, \vek{s})$ be the \emph{sequential encoding}
$\mathrm{LT}_\mathrm{SEQ}^{4,2}$ from~\cite{S05} simplified by
eliminating pure literals $\neg s_{1,2}$, $s_{3,1}$.
This is an encoding of the constraint at-most-$2$ on the variables
$\vek{x}=(x_1, x_2, x_3, x_4)$ with auxiliary variables $\vek{s}$.
Namely, $\theta'(\vek{x}, \vek{s})$ is a Horn formula consisting of the clauses
$$
\begin{array}{lll}
(\neg x_1 \vee s_{1,1}), & (\neg x_2 \vee \neg s_{1,1} \vee s_{2,2}), & (\neg x_3 \vee \neg s_{2,2}),\\
(\neg s_{1,1} \vee s_{2,1}), & (\neg s_{2,2} \vee s_{3,2}),\\
(\neg x_2 \vee s_{2,1}), & (\neg x_3 \vee \neg s_{2,1} \vee s_{3,2}), & (\neg x_4 \vee \neg s_{3,2})\,.
\end{array}
$$
One can verify that \(\theta'\) is propagation complete.
This follows by the results of~\cite{BM12} from the fact that
all prime implicates of \(\theta'\) can be
derived by a series of non-merge resolutions.
The same argument can be used to show that $\mathrm{LT}_\mathrm{SEQ}^{n,k}$
is PC for every $n \ge 2$ and $n \ge k \ge 1$.

The formula $\phi(\vek{x}) \wedge \theta(\vek{x})$ consists of all the
prime implicates of the exactly-$2$ constraint, so it is a propagation complete
representation of this constraint. Using unit propagation and
resolution, one can verify that the formula
$\phi(\vek{x}) \wedge \theta'(\vek{x}, \vek{s}) \wedge \neg s_{3,2} \wedge \neg x_4$
implies $\phi(\vek{x}) \wedge (\neg x_1 \vee \neg x_2) \wedge (\neg x_1 \vee \neg x_3)
\wedge (\neg x_2 \vee \neg x_3) \wedge \neg x_4$, so it is
contradictory. On the other hand, unit propagation does not derive a contradiction.
It follows that $\phi(\vek{x}) \wedge \theta'(\vek{x}, \vek{s})$
is an encoding of exactly-$2$ constraint on the variables $\vek{x}$
which is not unit refutation complete.

In the next subsections, we consider the at-most-one and
the exactly-one constraints as $\theta$ and describe their encodings
$\theta'$ such that the replacement of $\theta$ by $\theta'$ in an arbitrary
encoding preserves a specified propagation strength.

\subsection{At-most-one Constraint}
\label{sub:amo}

Recall that $\amo{A}$ is the at-most-one constraint for a set
of literals $A$ and by \(\amorep{A}\) we denote the canonical
representation of \(\amo{A}\) consisting of all the prime
implicates of this function. For simplicity,
let us assume that all the literals in $A$ are positive and form a
vector of the variables $\vek{x}$.

The number of clauses in $\amorep{\vek{x}}$
is quadratic, however, there are linear size encodings using auxiliary
variables~\cite{S05,C10,FG10,FPDN05,HN13,NM15}. Most of the encodings of the
at-most-one constraint described in the literature are prime 2-CNFs, hence,
they are propagation complete by~\cite{BBCGK13}. We show that using these
encodings in place of \(\amorep{\vek{x}}\) inside another
encoding preserves unit refutation completeness of the whole encoding.

The proof of the result uses the assumption that an encoding
$\amoenc{\vek{x}, \vek{z}}$ does not
contain positive ocurrences of the variables from $\vek{x}$. This is
a natural assumption satisfied by any irredundant 2-CNF encoding of
$\amo{\vek{x}}$, in particular, by most of the encodings suggested in the
references above.

\begin{proposition}
   \label{lem:embed-amo}
Assume that \(\varphi(\vek{x}, \vek{x}', \vek{y})\land\amorep{\vek{x}}\)
is a URC encoding of a constraint $f(\vek{x}, \vek{x}')$.
Let \(\amoenc{\vek{x}, \vek{z}}\) be
   a PC encoding of \(\amo{\vek{x}}\) with auxiliary variables
   \(\vek{z}\) which does not contain positive literals on \(\vek{x}\).
Then \(\varphi(\vek{x}, \vek{x}', \vek{y})\land\amoenc{\vek{x}, \vek{z}}\)
is a URC encoding of the constraint \(f(\vek{x}, \vek{x}')\).
\end{proposition}

\begin{proof}
Using~\eqref{eq:general-embedding},
we obtain that \(\varphi\land\amoenc{\vek{x}, \vek{z}}\)
is an encoding of \(f(\vek{x}, \vek{x}')\) with auxiliary
variables $\vek{y} \cup \vek{z}$.
Let us show that it is a URC encoding.

Let $\alpha = \alpha_x \cup \alpha_{x', y} \cup \alpha_z$, where
$\alpha_x \subseteq \lit{\vek{x}}$,
$\alpha_{x',y} \subseteq \lit{\vek{x}' \cup \vek{y}}$,
$\alpha_z \subseteq \lit{\vek{z}}$.
Our goal is to show that if unit propagation does not derive a contradiction
from the formula \(\varphi\land\amoenc{\vek{x}, \vek{z}} \land
\alpha_x \wedge \alpha_{x',y} \wedge \alpha_z\),
then this formula is satisfiable. So, assume
\begin{equation} \label{eq:amo-no-contradiction}
\varphi\land\amoenc{\vek{x}, \vek{z}} \land
\alpha_x \wedge \alpha_{x',y} \wedge \alpha_z \not\vdash_1 \bot \,.
\end{equation}
Without loss of generality, we can assume that \(\alpha\) is closed under unit propagation
in \(\varphi\land\amoenc{\vek{x}, \vek{z}}\), since replacing $\alpha$ with its
closure does not affect satisfiability of the formula. Let us prove
$$
\varphi\land\amorep{\vek{x}}\land \alpha_x\land\alpha_{x',y}\not\vdash_1\bot
$$
by contradiction.
If \(\varphi\land\amorep{\vek{x}}\land\alpha_x\land\alpha_{x',y}\vdash_1\bot\),
then also \(\varphi\land\amoenc{\vek{x}, \vek{z}}\land\alpha_x\land\alpha_{x',y}\vdash_1\bot\),
since \(\amoenc{\vek{x}, \vek{z}}\) is a PC encoding, but this
contradicts~\eqref{eq:amo-no-contradiction}.
Since \(\varphi\land\amorep{\vek{x}}\) is a URC encoding, we get that
\(\varphi\land\amorep{\vek{x}}\land\alpha_x\land\alpha_{x',y}\) has a
   satisfying assignment \(\assign{a}_x\cup\assign{a}_{x',y}\), where
   \(\assign{a}_x\) and \(\assign{a}_{x',y}\) denote the parts of the
   satisfying assignment on the variables in \(\vek{x}\) and
\(\vek{x}' \cup \vek{y}\) respectively.

It remains to show that the formula
$\amoenc{\vek{x}, \vek{z}}\land \assign{a}_x \land\alpha_z$
is satisfiable by a suitable assignment of the variables $\vek{z}$.
If $\assign{a}_x$ contains only negative literals,
then the formulas
$\amoenc{\vek{x}, \vek{z}}\land \assign{a}_x \land\alpha_z$ and
$\amoenc{\vek{x}, \vek{z}} \land\alpha_z$ are equisatisfiable,
since each clause of $\amoencN$ containing a literal on a variable
from $\vek{x}$ is
satisfied by $\assign{a}_x$ by the assumption that the variables $\vek{x}$ occur
only negatively. Equivalently, $\assign{a}_x$ is an autarky.
Since $\amoencN$ is PC, $\amoenc{\vek{x}, \vek{z}} \land\alpha_z$ is satisfiable
by~\eqref{eq:amo-no-contradiction} and the proof
is finished in this case.
If $\assign{a}_x$ contains a positive literal $x$, we have
$\neg x \not\in \alpha_x$, since $\assign{a}_x$ is a satisfying
assignment of a formula containing $\alpha_x$.
Since the partial assignment $\alpha_x\land\alpha_z$
is closed under the unit propagation in $\amoenc{\vek{x}, \vek{z}}$,
we have that neither \(\amoenc{\vek{x},
      \vek{z}}\land\alpha_x\land\alpha_z\not\vdash_1\neg x\), nor
\(\amoenc{\vek{x},
      \vek{z}}\land\alpha_x\land\alpha_z\not\vdash_1\bot\). Since
$\amoencN$ is PC, we have that
\(\amoenc{\vek{x},
      \vek{z}}\land\alpha_x\land\alpha_z\not\models\neg x\).
   The formula
   $\amoenc{\vek{x}, \vek{z}}\land\alpha_x\land\alpha_z\wedge x$ is thus satisfiable.
Moreover, this formula has a satisfying assignment consistent with
$\assign{a}_x$ since the literal $x$ forces negative literals on all the
other variables of $\amo{\vek{x}}$.
\end{proof}

\subsection{Exactly-one Constraint}
\label{ssec:exactly-one}

Recall that $\exone{A}$ is the exactly-one constraint on a set of
literals $A$ and by \(\exonerep{A}\) we denote the canonical
representation of \(\exone{A}\) consisting of all the prime
implicates of this function. For simplicity,
let us assume that all the literals in $A$ are positive and form
a vector of the variables $\vek{x}=(x_1,\ldots,x_n)$.

The exactly-one constraint \(\exone{\vek{x}}\) can be represented
as \(\amo{\vek{x}}\) together with the clause
\(x_1\lor\dots\lor x_n\). Let us demonstrate that the encoding
obtained in this way is not in general propagation complete.
For example, consider the encoding
\begin{align*}
   \phi=&(\neg x_1 \lor \neg x_2)(\neg x_1 \lor s_2)
   (\neg x_2\lor s_2)(\neg s_2 \lor \neg x_3)(\neg s_2\lor \neg x_4)
   (\neg x_3\lor \neg x_4)\\
   &(x_1\lor x_2\lor x_3\lor x_4)\,.
\end{align*}
of \(\exone{x_1, x_2, x_3, x_4}\) obtained, if $\amo{x_1, x_2, x_3, x_4}$
is represented by the sequential encoding
$\mathrm{LT}_\mathrm{SEQ}^{4,1}$ from~\cite{S05}
(called AMO sequential counter encoding in~\cite{NM15})
simplified by eliminating $s_1$ and $s_3$ by DP (Davis-Putnam) resolution.
Assume the partial assignment \(\neg x_3\land\neg x_4\). We have
\(\phi \land \neg x_3\land\neg x_4\models s_2\)
and the unit propagation does not derive \(s_2\),
so the encoding is not propagation complete.
On the other hand, the encoding is URC, since it is obtained from
a PC formula by adding a single clause.

Let us recall the ladder encoding~\cite{GN04} of $\exone{x_1,\ldots,x_n}$
which uses auxiliary variables $z_1, \ldots, z_{n-1}$. For simplicity,
let $z_0=1$ and $z_n=0$ be constants. The ladder encoding
can then be obtained by an expansion of
\begin{equation} \label{eq:ladder}
\epsilon_n(x_1, \dots, x_n, \vek{z}) =
\bigwedge_{i=2}^{n-1} (z_{i-1} \vee \neg z_i) \wedge
\bigwedge_{i=1}^{n} (x_i \Leftrightarrow z_{i-1} \wedge \neg z_i)
\end{equation}
into clauses. The resulting formula can be expressed as
\begin{equation} \label{eq:ladder-from-eo}
\epsilon_n(x_1, \dots, x_n, \vek{z}) =
\bigwedge_{i=1}^{n} \exonerep{\neg z_{i-1}, x_i, z_i}
\end{equation}
which implies that $\epsilon_n$ is a conjunction
of a sequence of $n$ PC formulas $\exonerep{\neg z_{i-1}, x_i, z_i}$
where each two consecutive ones share a single variable. It follows
that the ladder encoding is PC, see the proof of Proposition 5
in~\cite{BM12}.
Moreover, the \(\exone{\vek{x}}\) constraint satisfies
\begin{equation}\label{eq:eo-eliminate}
\exone{x_1, \ldots, x_n} \Leftrightarrow
(\exists z)
[\exone{x_1, \ldots, x_j, z} \wedge
\exone{\neg z, x_{j+1}, \ldots, x_n}]
\end{equation}
where $z$ is an auxiliary variable and $1 \le j \le n-1$. If we fix an
index \(i\in\{1, \dots, n-1\}\), we can use~\eqref{eq:eo-eliminate} to
eliminate all auxiliary variables \(z_j\), \(j\neq i\) which gives us
\begin{equation} \label{eq:decompose-eo}
\epsilon_n(\vek{x}, \vek{z}) \models
\exone{x_1, \ldots, x_{i}, z_i} \wedge
\exone{\neg z_i, x_{i+1}, \ldots, x_n}\,.
\end{equation}

In order to show that replacing $\exonerep{\vek{x}}$ with
the ladder encoding inside a larger encoding preserves
propagation completeness, we prove a more general statement.
The ladder encoding satisfies its assumption, since it is
propagation complete and using~\eqref{eq:decompose-eo}
we have
$$
\epsilon_n(\vek{x}, \vek{z}) \models
(z_i \Leftrightarrow \neg x_1 \wedge \cdots \wedge \neg x_i)
$$
and
$$
\epsilon_n(\vek{x}, \vek{z}) \models
(\neg z_i \Leftrightarrow \neg x_{i+1} \wedge \cdots \wedge \neg x_n)
$$
for every $i=1,\ldots,n-1$.

\begin{proposition}
   \label{lem:embed-eo}
Assume that \(\varphi(\vek{x}, \vek{x}', \vek{y})\land\exonerep{\vek{x}}\)
is a PC encoding of a constraint $f(\vek{x}, \vek{x}')$.
Let \(\exoneenc{\vek{x}, \vek{z}}\) be
   a PC encoding of \(\exone{\vek{x}}\) with auxiliary variables
   \(\vek{z}\) which satisfies the following property: for every
   literal \(l\in\lit{\vek{z}}\) there is a partial assignment
   \(h(l)\subseteq\lit{\vek{x}}\), such that
\(\exoneenc{\vek{x}, \vek{z}} \models (l \Leftrightarrow h(l))\).
   Then \(\varphi(\vek{x}, \vek{x}', \vek{y})\land\exoneenc{\vek{x}, \vek{z}}\)
   is a PC encoding of the constraint $f(\vek{x}, \vek{x}')$.
\end{proposition}

\begin{proof}
Using~\eqref{eq:general-embedding}, we obtain that
\(\varphi\land\exoneenc{\vek{x}, \vek{z}}\)
is a CNF encoding of \(f(\vek{x}, \vek{x}')\).
Moreover, the assumption implies that the formula
$\phi(\vek{x}, \vek{x}', \vek{y}) \wedge \exoneenc{\vek{x}, \vek{z}}$
satisfies the assumption of Proposition~\ref{prop:DC-encoding} with
$\vek{x}'$ replaced with $\vek{x}' \cup \vek{y}$,
$\vek{y}$ replaced with an empty set of variables,
$\theta(\vek{x})$ replaced with $\exonerep{\vek{x}}$, and
$\theta'(\vek{x}, \vek{z})$ replaced with $\exoneenc{\vek{x}, \vek{z}}$.
It follows that
$\phi(\vek{x}, \vek{x}', \vek{y}) \wedge \exoneenc{\vek{x}, \vek{z}}$
is PC on the variables $\vek{x} \cup \vek{x}' \cup \vek{y}$.

In order to show that $\phi \wedge \exoneenc{\vek{x}, \vek{z}}$
is PC on all its variables, consider a partial assignment
$\alpha = \alpha_x \cup \alpha_{x',y} \cup \alpha_z$ where
$\alpha_x \subseteq \lit{\vek{x}}$,
$\alpha_{x',y} \subseteq \lit{\vek{x}' \cup \vek{y}}$,
$\alpha_z \subseteq \lit{\vek{z}}$ and a
   literal \(l\in\lit{\vek{x} \cup \vek{x}' \cup\vek{y}\cup\vek{z}}\).
Without loss of generality, we can assume that \(\alpha\) is closed
under unit propagation in
   \(\varphi\land\exoneenc{\vek{x}, \vek{z}}\),
since replacing \(\alpha\) with its closure
does not affect validity of the conditions~\eqref{eq:pc-enc-1}
and~\eqref{eq:pc-enc-2}.
In order to show that the implication from~\eqref{eq:pc-enc-1}
to~\eqref{eq:pc-enc-2} holds also in this case, assume
   \begin{equation}
      \label{eq:embed-eo:1}
      \varphi\land\exoneenc{\vek{x}, \vek{z}}
      \land \alpha_x \wedge \alpha_{x',y} \wedge \alpha_z \models l\text{.}
   \end{equation}
Our goal is to show that if
$$
\varphi\land\exoneenc{\vek{x}, \vek{z}}\land
\alpha_x \wedge \alpha_{x',y} \wedge \alpha_z \not\vdash_1\bot
$$
then
   \begin{equation}
      \label{eq:embed-eo:2}
      \varphi\land\exoneenc{\vek{x}, \vek{z}}
      \land\alpha_x \wedge \alpha_{x',y} \wedge \alpha_z \vdash_1 l\text{.}
   \end{equation}
Let $h(\alpha_z) \subseteq \lit{\vek{x}}$ denote the partial assignment
consisting of $h(g)$ for all $g \in \alpha_z$. Since $\exoneencN$ is PC,
unit propagation on $\exoneenc{\vek{x}, \vek{z}} \wedge \alpha_z$
derives all the literals in $h(\alpha_z)$ and
unit propagation on $\exoneenc{\vek{x}, \vek{z}} \wedge h(\alpha_z)$
derives all the literals in $\alpha_z$. Since $\alpha_x \cup \alpha_z$
is closed under unit propagation in $\exoneencN$, we have
$h(\alpha_z) \subseteq \alpha_x$.
It follows that
$\exoneenc{\vek{x}, \vek{z}} \wedge \alpha_x \models \alpha_z$
and, consequently,
$$
\varphi\land\exoneenc{\vek{x}, \vek{z}}\land
\alpha_x \wedge \alpha_{x',y} \models l\text{\,.}
$$
If $l \in \lit{\vek{x} \cup \vek{x}' \cup \vek{y}}$, we get~\eqref{eq:embed-eo:2},
since $\phi \wedge \exoneenc{\vek{x}, \vek{z}}$ is
PC on the variables $\vek{x} \cup \vek{x}' \cup \vek{y}$.
If $l \in \lit{\vek{z}}$, then
$\exoneenc{\vek{x}, \vek{z}} \wedge l \models h(l)$ implies
$$
\varphi\land\exoneenc{\vek{x}, \vek{z}}\land
\alpha_x \wedge \alpha_{x',y} \models h(l)
$$
and we obtain
$$
\varphi\land\exoneenc{\vek{x}, \vek{z}}
\land\alpha_x \wedge \alpha_{x',y} \vdash_1 g
$$
for every literal $g \in h(l)$ again by
propagation on the variables $\vek{x} \cup \vek{x}' \cup \vek{y}$.
Together with the fact that $\exoneencN$ is
a PC encoding, we obtain $\exoneenc{\vek{x}, \vek{z}} \wedge h(l) \vdash_1 l$
and, hence,~\eqref{eq:embed-eo:2} as required.
\end{proof}

Let us point out that the assumption of Proposition~\ref{lem:embed-eo}
is satisfied also by an encoding $\varepsilon_n'$ which is slightly smaller
than the ladder encoding.
For \(n\le 4\), $\varepsilon_n'=\exonerep{x_1, \ldots, x_n}$.
If $n \ge 5$ is even, the encoding contains $n/2-2$ auxiliary variables
and has the form
\begin{equation} \label{eq:exone-explicit}
\begin{array}{rl}
\epsilon_n'(x_1, \dots, x_n, \vek{z}) = & \exonerep{x_1, x_2, x_3, z_1} \ \wedge\\
& \exonerep{\neg z_1, x_4, x_5, z_2} \ \wedge\\
& \cdots \\
& \exonerep{\neg z_{j-1}, x_{2j}, x_{2j+1}, z_j} \ \wedge\\
& \cdots \\
& \exonerep{\neg z_{n/2-3}, x_{n-4}, x_{n-3}, z_{n/2-2}} \ \wedge\\
& \exonerep{\neg z_{n/2-2}, x_{n-2}, x_{n-1}, x_n}
\end{array}
\end{equation}
If $n \ge 5$ is odd, the number of auxiliary variables is $n/2-3/2$ and the
encoding has the form~\eqref{eq:exone-explicit} except that the last
two constraints are
$$
\exonerep{\neg z_{n/2-5/2}, x_{n-3}, x_{n-2}, z_{n/2-3/2}} \wedge
\exonerep{\neg z_{n/2-3/2}, x_{n-1}, x_n}
$$
For both even and odd $n \ge 5$, the expansion~\eqref{eq:exone-explicit}
consists of at most $\frac{1}{2}\, n$ constraints $\exoneN$ of $3$ or $4$ variables
each of which consists of at most $7$ clauses. It follows that $\epsilon_n'$
has
$\frac{7}{2} \, n + O(1)$ clauses and at most $\frac{1}{2}\, n$ auxiliary variables
compared to $4n + O(1)$ clauses and $n-1$ auxiliary variables
of the ladder encoding.

\subsection{Application to the Encodings of MDNNF}
\label{sec:complexity}

In this section, we present upper bounds on the size of URC and PC encodings
of a smooth MDNNF, if the cardinality constraints
on the separators are expressed using the linear size encodings
instead of encodings of quadratic size with no auxiliary variables.
Assume, $\Ds$ is a smooth MDNNF and $\Dc$ is a smooth MDNNF that is
covered by separators obtained from $\Ds$ in step~\ref{enum:constr-separators}
of the proof of the main result.
Let $\psi_c$ and $\psi_p$ be the encodings obtained for
$\Dc$ in Section~\ref{sub:mdnnf}.

Assume \(S\) is a separator and denote \(\amoenc{S, \vek{z}_S}\) any
irredundant 2-CNF encoding of \(\amo{S}\) of linear size with auxiliary
variables \(\vek{z}_S\),
for example, the sequential counter encoding~\cite{S05,NM15}.
Recall that every such encoding satisfies the
assumption of Proposition~\ref{lem:embed-amo}. Similarly, let
\(\exoneenc{S, \vek{z}_S}\) denote a linear size encoding of
\(\exone{S}\) with auxiliary variables \(\vek{z}_S\) which satisfies
the assumption of Proposition~\ref{lem:embed-eo}, for example,
the ladder encoding discussed in Section~\ref{ssec:exactly-one}.

Consider an encoding \(\psi_c'(\dvar{\vek{x}}, \vek{y}, \vek{z})\)
obtained from \(\psi_c\) by using encoding \(\amoenc{S, \vek{z}_S}\)
instead of \(\amorep{S}\) for each
separator $S$ in the cover. In addition to the main variables
\(\dvar{\vek{x}}\) and auxiliary variables \(\vek{y}\), the encoding
\(\psi_c'\) uses auxiliary variables \(\vek{z}\) which is the union of
all sets \(\vek{z}_S\) for all separators.

Similarly, consider encoding \(\psi_p'(\dvar{\vek{x}}, \vek{y}, \vek{z})\)
which differs from \(\psi_p\) by using encoding
\(\exoneenc{S, \vek{z}_S}\) instead of
\(\exonerep{S}\) for each separator $S$ in the cover.
In addition to the main variables
\(\dvar{\vek{x}}\) and auxiliary variables \(\vek{y}\), the encoding
\(\psi_p'\) uses auxiliary variables \(\vek{z}\) which is the union of
all sets \(\vek{z}_S\) for all separators.

Encodings $\psi_c'$ and $\psi_p'$ are encodings of $\Ds$. Moreover,
encoding \(\psi_c'\) is URC by Proposition~\ref{lem:embed-amo} and
encoding \(\psi_p'\) is PC by Proposition~\ref{lem:embed-eo}.
We assume the the separator cover is constructed using the procedure
from Proposition~\ref{prop:separators-by-Level}. Hence,
the main terms in the upper bound on the size of $\psi_c'$ and $\psi_p'$
are the size (vertices and edges together)
of $\Ds$ and the total size $t$ of the separators in $\Dc$
estimated in Proposition~\ref{prop:separators-complexity}. By \emph{length
of an encoding} we mean the sum of the sizes of its clauses.

In the following estimate, we do not include direct encoding constraints
to the size. One of the reasons is that they can be shared among several
constraints in the instance, so they do not contribute to the size of each
of them. On the other hand, if the encoding is used separately, then it
is sufficient to consider these constraints after propagating
the clauses~\ref{N4}. The estimate of the size of the separator cover
from Proposition~\ref{prop:separators-complexity}
is valid for a cover which contains
$L_i$ for each $i=1,\ldots,n$ and the clauses $\exonerep{L_i}$ are
exactly the direct encoding constraints for $\dvar{x_i}$ after
propagating~\ref{N4}. As a consequence, we can assume that the
direct encoding constraints can be replaced by the encodings
$\exoneenc{L_i, \vek{z}_{L_i}}$. The encoding $\psi_p'$ contains this
encoding, so its size is included in the size estimate presented below.
The encoding $\psi_c'$ contains only $\amoenc{L_i, \vek{z}_{L_i}}$,
so replacing it with $\exoneenc{L_i, \vek{z}_{L_i}}$ increases
the size, however, both these encodings have size $\Theta(|L_i|)$,
so the asymptotic estimate does not change also in this case.

\begin{theorem}
   \label{thm:enc-size}
   Denote
   \(d=\max_{i=1}|\dom{x_i}|\) the maximum size of a domain, \(s\)
   the number of nodes of \(\Ds\), \(e\) the number of
   edges of \(\Ds\) and \(t\) the sum of the sizes of different
separators constructed for $\Dc$.
   Then \(\psi_c'(\dvar{\vek{x}}, \vek{y}, \vek{z})\) and
   \(\psi_p'(\dvar{\vek{x}}, \vek{y}, \vek{z})\) have
   \(O(nd+t)\) variables, \(O(nd+e+t)\) clauses, and length
   \(O(nd+e+t)\).
\end{theorem}
\begin{proof}
   Observe that the number of
   variables in \(\vek{z}\) is proportional to the total size of
   separators \(t\). Since \(|\dvar{\vek{x}}|\leq nd\) and \(|\vek{y}|\leq t\),
   we get that the number of
   variables in the encodings is bounded by \(O(nd+t)\).

   The total number of clauses in groups~\ref{N1} to~\ref{N3} is \(O(e)\).
   The number of clauses in group~\ref{N4} is at most \(nd\) and the
   total number of clauses in the encodings of cardinality constraints
   \(\amoenc{S, \vek{z}_S}\) and \(\exoneenc{S, \vek{z}_S}\) respectively
   is proportional to the total size of separators \(t\). Together,
   the number of clauses in the encodings is bounded by \(O(nd+e+t)\).

   Clauses of group~\ref{N1}
   have total length at most \(e\) and the same holds for the clauses
   of group~\ref{N3}. The rest of the clauses of both \(\psi_c'\) and
   \(\psi_p'\) have constant size and the length of both encodings is
   thus \(O(nd+e+t)\).
\end{proof}

\section{Conclusion and an Open Problem}
\label{sec:conclusion}

We demonstrated a propagation complete encoding for a smooth DNNF,
for which the previously known encodings implement only the domain
consistency. In this context, it is natural to ask the following.

\bigskip
\noindent\textbf{Question 1.} Assume, $\phi(\vek{x}, \vek{y})$ is
an encoding of a boolean function $f(\vek{x})$ with auxiliary
variables $\vek{y}$ that implements domain consistency.
Does this imply that there is an encoding $\phi'(\vek{x}, \vek{z})$ of
the same function of size polynomial in the size of $\phi$ with possibly
a different set of auxiliary variables $\vek{z}$ that is unit
refutation complete?

\bigskip
The results of Section~\ref{sec:amo-eo} for the at-most-one and
exactly-one constraints use quite specific properties of these
constraints. So, one can expect a negative answer to the following
question, however, a provable answer would be useful.

\bigskip
\noindent\textbf{Question 2.} Let $\theta(\vek{x})$ be the prime
CNF representation of the constraint at-most-$k$, where $k$ is a constant.
Is there a PC encoding $\theta'(\vek{x}, \vek{z})$ of $\theta(\vek{x})$
of linear size such that for every URC formula of the
form $\phi(\vek{x}, \vek{x}', \vek{y}) \wedge \theta(\vek{x})$
also the formula
$\phi(\vek{x}, \vek{x}', \vek{y}) \wedge \theta'(\vek{x}, \vek{z})$
is URC?

\paragraph{\textup{\textbf{Acknowledgements.}}}
Both authors gratefully acknowledge the
support by Grant Agency of the Czech Republic (grant No.~GA19--19463S).

\bibliography{ms}{}
\bibliographystyle{plain}

\end{document}